\pdfoutput=1

\documentclass[11pt]{article}

\usepackage[final]{acl}

\usepackage{times}
\usepackage{latexsym}

\usepackage[T1]{fontenc}

\usepackage[utf8]{inputenc}

\usepackage{microtype}

\usepackage{inconsolata}

\usepackage{graphicx}
\usepackage{amsmath}
\usepackage{amssymb}
\usepackage{booktabs} 
\usepackage{subfig}
\usepackage{multirow}
\usepackage{colortbl}

\title{\textbf{PTQ\textit{1.61}}: Push the Real Limit of Extremely Low-Bit Post-Training Quantization Methods for Large Language Models}

\author{Jiaqi Zhao\textsuperscript{1}, Miao Zhang\textsuperscript{1}\thanks{Corresponding Author}, Ming Wang\textsuperscript{1}, Yuzhang Shang\textsuperscript{2}, Kaihao Zhang\textsuperscript{1}, Weili Guan\textsuperscript{1}, \\
        \bf{Yaowei Wang\textsuperscript{1}, Min Zhang\textsuperscript{1}}\\
        \textsuperscript{1} Harbin Institute of Technology (Shenzhen)\\
        \textsuperscript{2} Illinois Institute of Technology\\
        \texttt{jiaqizhao0455@outlook.com, 190110509@stu.hit.edu.cn}\\
        \texttt{\{zhangmiao, guanweili, wangyaowei, zhangmin2021\}@hit.edu.cn}\\
        \texttt{yshang4@hawk.iit.edu, super.khzhang@gmail.com}}

\begin{document}
\maketitle
\begin{abstract}
Large Language Models (LLMs) suffer severe performance degradation when facing extremely low-bit (sub 2-bit) quantization. Several existing sub 2-bit post-training quantization (PTQ) methods utilize a mix-precision scheme by leveraging an unstructured fine-grained mask to explicitly distinguish salient weights, while which introduces an extra 1-bit or more per weight. To explore the real limit of PTQ, we propose an extremely low-bit PTQ method called PTQ1.61, which enables weight quantization to 1.61-bit for the first time. Specifically, we first introduce a one-dimensional structured mask with negligibly additional 0.0002-bit per weight based on input activations from the perspective of reducing the upper bound of quantization error to allocate corresponding salient weight channels to 4-bit. For non-salient channels binarization, an efficient block-wise scaling factors optimization framework is then presented to take implicit row-wise correlations and angular biases into account. Different from prior works that concentrate on adjusting quantization methodologies, we further propose a novel paradigm called quantization preprocessing, where we argue that transforming the weight distribution of the pretrained model before quantization can alleviate the difficulty in per-channel extremely low-bit PTQ. Extensive experiments indicate our PTQ1.61 achieves state-of-the-art performance in extremely low-bit quantization. Codes are available at \url{https://github.com/zjq0455/PTQ1.61}.
\end{abstract}

\section{Introduction}

Large Language Models (LLMs) such as LLaMA \cite{touvron2023llama1,touvron2023llama2} and GPT \citep{brown2020language,ouyang2022training} have demonstrated remarkable success in various natural language processing tasks. However, their colossal numbers of parameters bring tremendous storage and inference overheads. 
To alleviate the challenge, numerous model compression methods have been proposed such as quantization \cite{liu2022pd,huang2019efficient}, pruning \citep{frantar2023sparsegpt,ma2023llm} and knowledge-distillation \citep{gou2021knowledge,tunstall2023zephyr}. Among these works, post-training quantization (PTQ) methods \citep{yuan2023rptq,wei2022outlier} have garnered particular attention for LLMs due to their computational efficiency compared to quantization-aware training (QAT) \citep{liu2023llm,esser2019learned} and other compression methods. Although maintaining nearly lossless performance at 4-bit or 8-bit, most existing state-of-the-art PTQ approaches fail when attempting to quantize weights to extremely low bit-width, \textit{i.e.}, 1-bit or sub 2-bit.

\begin{figure}[t]
  \centering
  \includegraphics[width=2.4in]{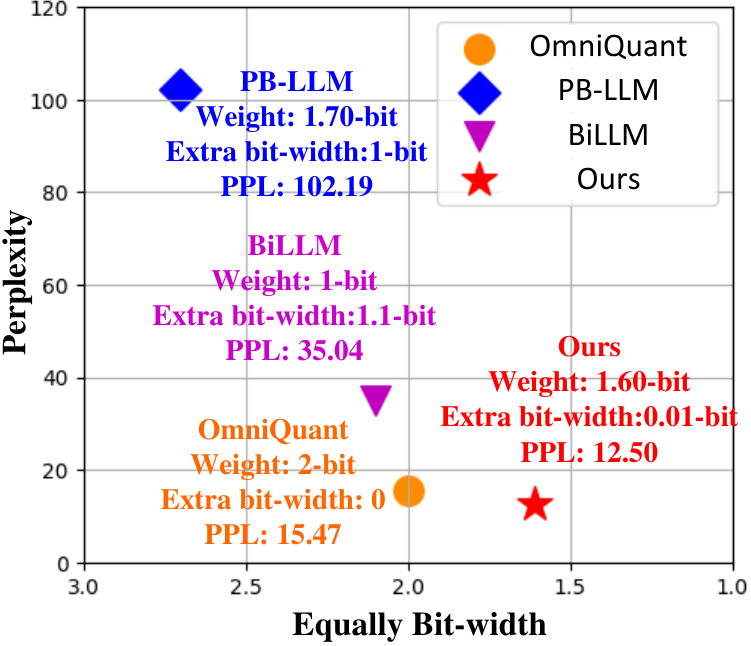}
  \caption{An overview of performance (Perplexity on WikiText2) and bit-width achieved by our \textbf{PTQ\textit{1.61}} and other extremely low-bit PTQ methods on LLaMA-7B.}
  \label{Fig1}
\end{figure}

PB-LLM \citep{shang2023pb} and BiLLM \citep{huang2024billm} are two most recent sub 2-bit PTQ methods for LLMs. They selectively preserve a portion of salient weights at 8-bit or with fine processing while quantizing the remaining weights to 1-bit. Although they demonstrate promising results, they are plagued by two critical issues. \textbf{\textit{Firstly and the most importantly}}, both methods introduce additional unstructured fine-grained masks to distinguish salient weights which requires additional 1-bit per weight to store the mask and leads the memory of the quantized model to exceeding 2-bit per weight, where PB-LLM with 2.7-bit and BiLLM with 2.1-bit respectively (see Appendix \ref{appendix_a}). \textbf{\textit{Secondly}}, they independently and analytically derive the row-wise scaling factors used for mitigating binarization magnitude errors \citep{rastegari2016xnor}, violating the fact that weights exhibit implicit row-wise dependencies \citep{clark2019does,vig2019analyzing} and angular biases \citep{lin2020rotated}. 

Motivated by issues above and to push the real limit of PTQ methods on extremely low-bit quantization, we propose an extremely low-bit (\textbf{1.61-bit}) PTQ method for LLMs called \textbf{PTQ\textit{1.61}}. Specifically, to eliminate the significant additional memory consumption caused by unstructured fine-grained masks, we dissect the quantization error through mathematical derivation to identify the structural influencing factors within it, and find that the upper bound of quantization error is significantly affected by input activation channels. Based on this discovery we propose a \textit{one-dimensional structured mask} to preserve corresponding salient channels in the weight matrix at 4-bit, and successfully reduce the extra bit-width for each weight from over 1-bit to a negligible extent (0.0002-bit). Additionally, in order to capture the implicit row-wise correlations and directional biases jointly, we introduce a novel \textit{efficient block-wise scaling factors optimization framework}. 

In addition, unlike previous studies which always take the pretrained model with the best performance as the starting point for quantization, we find that the weights distribution also immensely affects the quantization performance. Specifically, existing per-channel PTQ methods usually consider a row-wise quantization pattern that assigns the same quantization parameter to all weights in a channel, while the distribution of salient weights in the pretrained model is scattered, which leads to significant quantization errors. Motivated by this row-wise nature, we propose a novel \textit{preprocessing strategy} for LLMs quantization, which first transforms the weight distribution into a row-wise pattern through a lightweight restorative LoRA alignment, so that the preprocessed model is more suitable for per-channel PTQ than pretrained model. The proposed preprocessing strategy can be also applied to other extremely low-bit PTQ methods with notable performance enhancement, as shown in Figure \ref{bar-chart}. We further discuss the differences and advantages of our preprocessing strategy from existing post-quantization parameter-efficient fine-tuning (PEFT) approaches \citep{dettmers2023qlora} in Section \ref{preprocessing} and Appendix \ref{appendix_d}.

With these enhancements, \textbf{PTQ\textit{1.61}} effectively quantizes the weights to extremely low-bit with outstanding performance, as illustrated in Figure \ref{Fig1}. Our key contributions can be summarized as:
\begin{itemize}
\item To explore the real limitation of post-training quantization, we present an efficient extremely low-bit PTQ method for LLMs named \textbf{PTQ\textit{1.61}} which is the \textbf{\textit{first}} PTQ study to truly reduce the effective bit-width of LLMs weights to sub 2-bit (1.61-bit) with acceptable performance degradation.

\item Different from leveraging memory-intolerable unstructured masks to preserve salient information, we propose a one-dimensional structured mask based on input activations to reduce the upper bound of quantization errors, which only introduces negligible 0.0002-bit for each weight.

\item We further present a novel efficient block-wise optimization strategy to learn scaling factors to further consider the implicit row-wise dependencies and angular biases.

\item We demonstrate that pretrained model is not amenable to per-channel PTQ and accordingly propose a quantization preprocessing paradigm based on restorative LoRA to transform salient weights as a row-wise pattern to further enhance the quantization performance.
\end{itemize}

\begin{figure*}[t]
  \centering
  \includegraphics[width=5.6in]{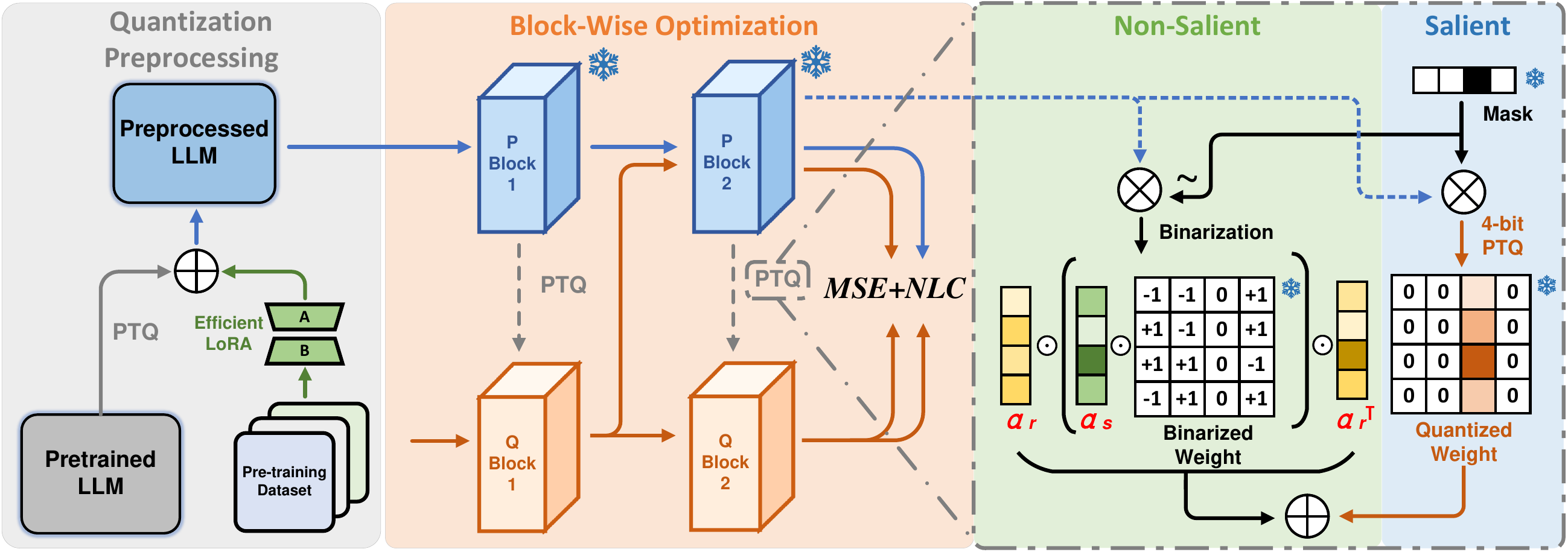}%
  \caption{An overview of our \textbf{PTQ\textit{1.61}}. Utilizing quantization preprocessing, the pretrained model is transformed into a row-wise pattern which is amenable to channel-wise quantization. Initially, structured masks are obtained to distinguish salient weights channels based on channel-wise magnitude of input activation. Salient weight channels undergo 4-bit quantization to retain crucial information, while non-salient weights are binarized with the aid of learnable scaling factors updated by novel block-wise optimization framework.}
  \label{PTQ1.61}
\end{figure*}

\section{Related Works}
\subsection{Post-Training Quantization}
Post-training quantization is an efficient and expeditious quantization approach which merely necessitates a limited amount of calibration data to statistically determine the quantization parameters that help to scale float values to low-bit. AdaRound \citep{nagel2020up} analyzes the quantization errors and employs a layer-wise optimization approach to learn the optimal rounding mechanism. BrecQ \citep{li2021brecq} divides the model weights into multiple blocks and independently quantizes each block, allowing for finer control over quantization errors.

In recent years, there has been a growing research interest in PTQ methods for LLMs. GPTQ \citep{frantar2022gptq} quantizes weights column-wise based on the Hessian matrix and dynamically updates the remaining weights to compensate for quantization errors. AWQ \citep{lin2023awq} retains 1\% of salient weights and calculates quantization parameters based on output activations. ZeroQuant \citep{yao2022zeroquant} performs group-wise quantization for weights and finer-grained per-channel quantization for activations. SmoothQuant \citep{xiao2023smoothquant} considers that the cause of quantization errors in activations lies in the presence of channel-wise outliers and proposes smoothing parameters to reduce their magnitude. OmniQuant \citep{OmniQuant} combines the advantages of previous works and learns the optimal smoothing and quantization parameters through back propagation, making it the current state-of-the-art PTQ method for LLMs.

Unfortunately, for extremely low-bit (sub 2-bit) quantization, which offers the highest compression ratio, the performance of such methods generally suffers significantly.

\subsection{Extremely Low-Bit Quantization}

Extremely low-bit quantization refers to approaches where the effective bit-width for weights is sub 2-bit. It has been widely welcomed due to significant compression ratio but suffers from severe performance degradation. BNN \citep{courbariaux2016binarized} is the first model binarization method and 
XNOR-Net \citep{rastegari2016xnor} presents scaling factors which reduce binarization errors with acceptable additional memory cost. RBNN \citep{lin2020rotated} indicates that except for magnitude gaps, angular biases ought to be considered so that extra rotation matrices are introduced to overcome the drawback.

Several extremely low-bit QAT methods for LLMs \citep{xu2024onebit,wang2023bitnet,ma2024era} have been proposed recently. Regrettably, the immense computational resource consumption and the lack of open-source availability have hindered their widespread application so that there is a growing demand for more economical PTQ methods. PB-LLM \citep{shang2023pb} investigate the importance of salient weights and design extra 1-bit unstructured masks to retain them into 8-bit while binarizing the others. BiLLM \citep{huang2024billm} further presents finer-grained masks to divide into multi-groups for binarization using different scaling factors. However, the fine-grained masks that cannot be compressed in both methods results in the equivalent bitwidths exceeding 2 bits. To make contributions for truly extremely low-bit PTQ research, we propose \textbf{PTQ\textit{1.61}} which addresses the issues above and obtains promising performance.

\section{PTQ\textit{1.61}}

In this section, we provide a detailed introduction to our \textbf{PTQ\textit{1.61}}, an extremely low-bit PTQ method for LLMs as demonstrated in Figure \ref{PTQ1.61}. We begin with briefly reviewing the basic concepts of model quantization and binarization in Section 3.1. Subsequently, to preserve salient information while avoiding insufferable memory overheads brought by unstructured masks in previous methods, we analyze the impact factors on quantization errors and then devise a one-dimensional mask based on input activations with negligible bit-width in Section 3.2. In Section 3.3, a novel block-wise optimization strategy is introduced for binarization to obtain optimal scaling factors considering implicit dependencies and angular biases. In Section 3.4, we explain why the pretrained model is not suitable for per-channel PTQ and how our proposed quantization preprocessing strategy works.

\subsection{Preliminaries}
\label{preliminaries}
\paragraph{Model Quantization}

Model quantization aims to convert float weights into corresponding low-bit integer forms thereby reducing computational and memory overheads. The quantization function can be elaborated as:
\begin{equation}
\mbox{W}_{q} = \mbox{clamp}(\lfloor\frac{\mbox{W}}{S_{q}}\rceil+Z_{q},0,2^{b}-1),
\label{higher-bit}
\end{equation}
where $\mbox{W} \in \mathbb{R}^{n \times m}$ and $\mbox{W}_{q} \in \mathbb{R}^{n \times m}$ indicate full-precision and quantized weights respectively. $\lfloor \cdot \rceil$ denotes round-to-nearest operator. $S_{q}$ is the quantization scalar and $Z_{q}$ represents zero-point. 

\paragraph{Model Binarization}

Binarization represents the most extreme form of quantization (1-bit) where weights are assigned as $\pm 1$ determined by the sign function. In more details:
\begin{equation}
\text{sign}(\mbox{W}) = \begin{cases}
+1, & \mbox{W} \geq 0, \\
-1, & \mbox{W}<0. \\
\end{cases}
,\quad \mbox{W}_{b} = \alpha \text{sign}(\mbox{W}),
\label{bi-eq}
\end{equation}
where $\alpha$ denotes scaling factors commonly used in previous methods \citep{bulat2019xnor,xu2021recu,liu2018bi,shang2023pb,huang2024billm} to reduce binarization errors and $\mbox{W}_{b} \in \mathbb{R}^{n \times m}$ is binarized weights with scaling factor $\alpha$. Assuming that weights in each row of $\mbox{W}$ are independent, we define $w$ as a row of weights with $n_{w}$ elements. The corresponding scaling factor $\alpha_w$ can be derived analytically by $\alpha_w = \frac{\lVert w \rVert_1}{n_{w}}$. 

\subsection{Structured Mask}
\label{structured-mask}

Following \citep{shang2023pb} and \citep{huang2024billm}, we recognize that partially preserving higher-bit salient information is crucial for reducing quantization errors. However, their unstructured masks cannot be compressed, resulting in additional memory overheads due to the scattered nature of salient weights within the weight matrix. Hence, our objective is to identify factors that significantly impact quantization errors while maintaining some degree of regularity.

\begin{figure}[t]
\centering
\subfloat{\includegraphics[width=1.05in]{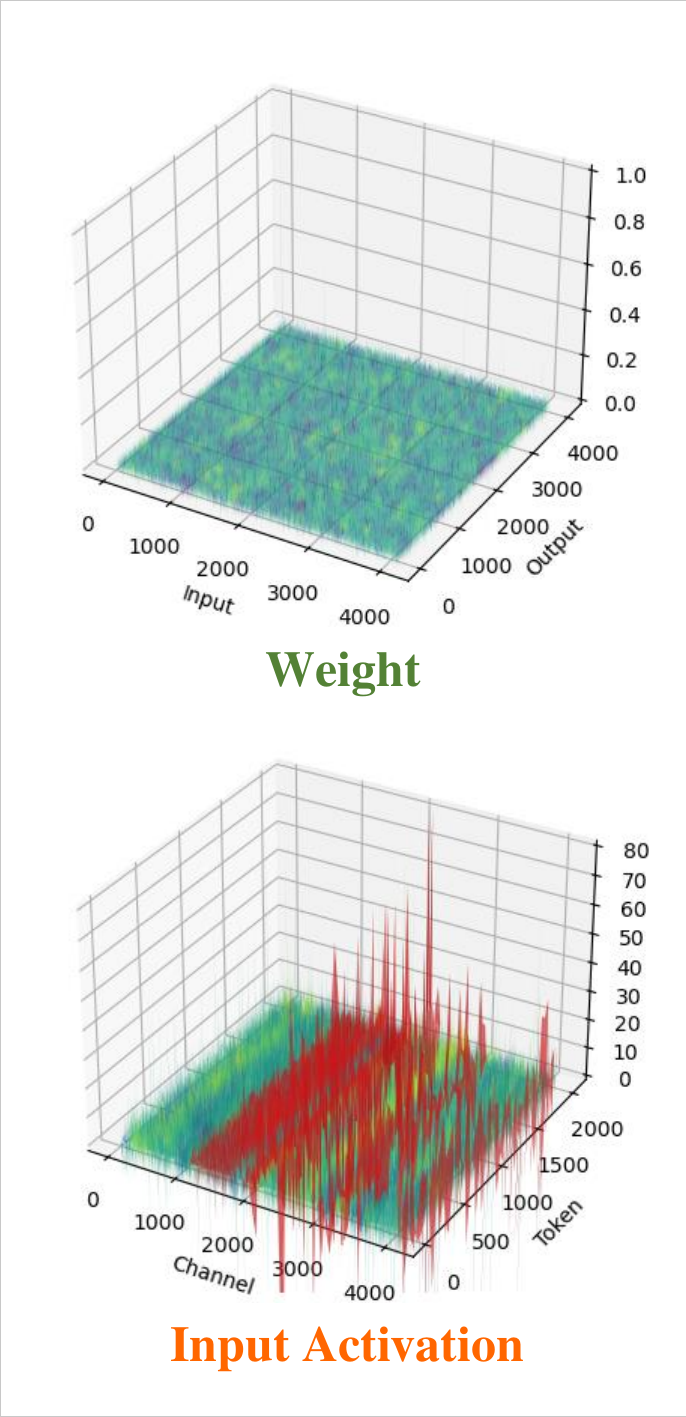}%
\label{magnitude-comparison}}
\subfloat{\includegraphics[width=1.99in]{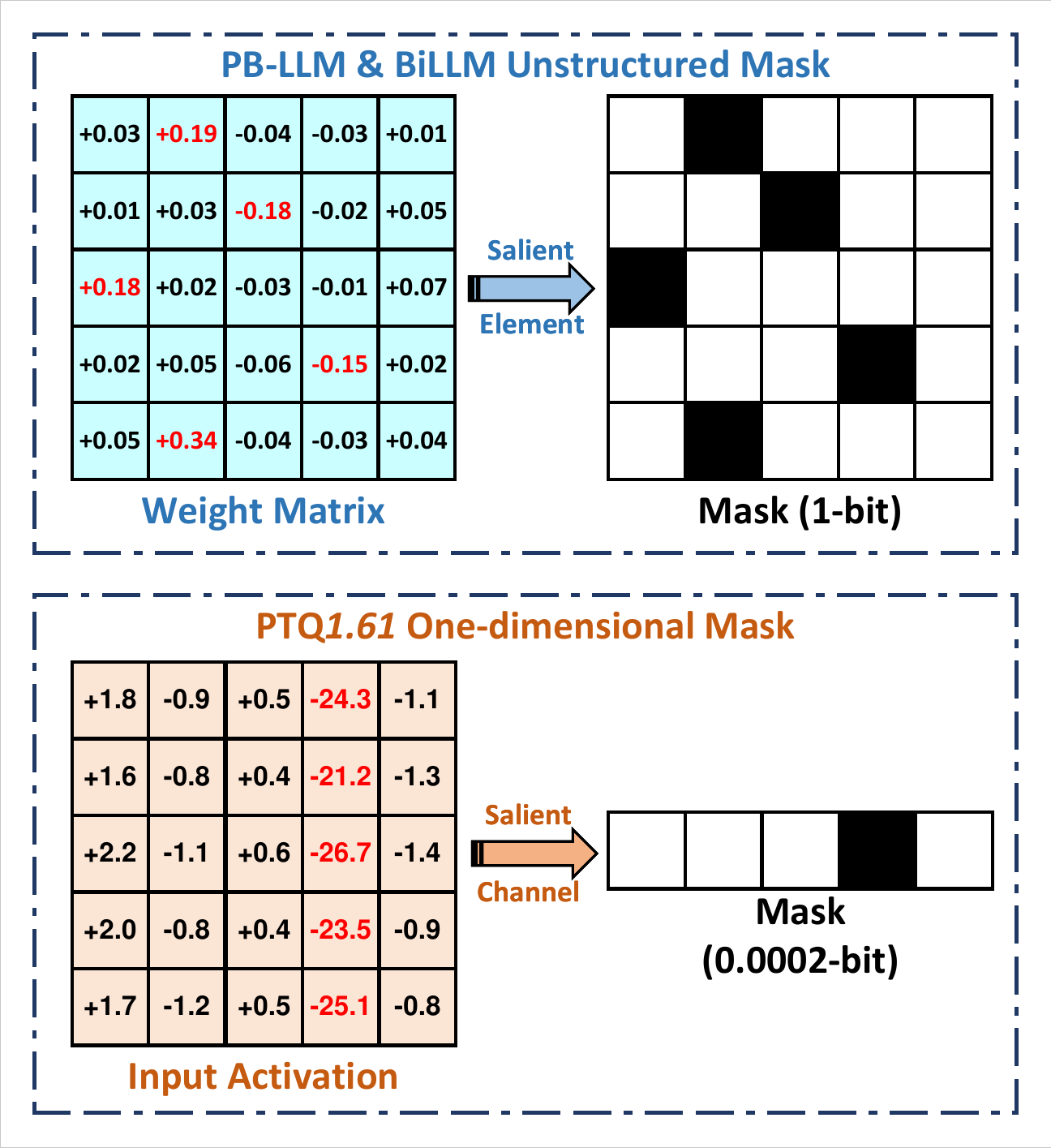}%
\label{mask comparison}}
\caption{{\bf{(a)}} The magnitude and distribution comparison between input activations and weights; {\bf{(b)}} Previous masks are unstructured and uncompressible. In contrast our mask is one-dimensional with only 0.0002-bit.}
\label{mask}
\end{figure}

For each quantized layer, its quantization error is expressed as the gap relative to full-precision output. This can be restated as:
\begin{equation}
\mathcal{E} = | \mbox{X} \mbox{W}_{q}^{T} -\mbox{X} \mbox{W}^{T} | = | \mbox{X}( \mbox{W}_{q}^{T} -\mbox{W}^{T}) |,
\label{E}
\end{equation}
where $\mathcal{E}$ is the layer-wise quantization error and $\mbox{X} \in \mathbb{R}^{t\times m}$ indicates the input activation of the layer. To further explore hidden key factors, we give the visualization of $\mbox{X}$ and $\mbox{W}$ as Figure \ref{magnitude-comparison} from which we can observe that in contrast to the chaotic distribution of weights, input activations exhibit a clear channel-wise pattern, where the variance of each token within a channel is very small. Therefore, we re-define $\mbox{X}$ as a column-wise set $\mbox{X} = \{x_1,x_1,...,x_m | x_1 \in \mathbb{R}^{n} \}$ so that Equation \eqref{E} will be modified as:

\begin{equation}
\begin{aligned}
\mathcal{E} = | \sum_{i=1}^{m} x_i(w_{i,j}^{q}-w_{i,j})_{j=1,...,n} | \\
\leq  \sum_{i=1}^{m} (|x_i| \sum_{j=1}^{n}| w_{i,j}^{q}-w_{i,j}|),
\label{E_2}
\end{aligned}
\end{equation}
where $w_{i,j}$ denotes the element in position $(i,j)$ of $\mbox{W}^{T}$. As demonstrated in Equation \eqref{E_2}, \textit{the upper bound of quantization error is related to the magnitude of input activations for the $i$-th channel and weight matrix for the $i$-th row.} Figure \ref{magnitude-comparison} highlights that the magnitude of activations is about 1000 times larger than that of weights especially for top-20\% channels. Therefore, we propose a one-dimensional mask to save the $i$-th row of $\mbox{W}$ at 4-bit, maintaining salient information in channel $i$ of input activations to reduce the upper bound of quantization error as shown in Figure \ref{mask comparison}. This improvement successfully reduces the additional bit-width introduced by masks to 0.0002-bit (see Appendix) and limit the equally weight bit-width at sub 2-bit (1.61-bit) for the first time among all existing PTQ methods. The light blue area of Figure \ref{PTQ1.61} illustrates the role of our structured mask in the entire quantization process. Notably, OWQ \citep{lee2024owq} and AWQ \citep{lin2023awq} also take into account the relationship between input activation and weight, but our motivation and performance are entirely different. We have elaborated on this part in Appendix \ref{appendix_b}.

\subsection{Block-wise Scaling Factors Optimization}
For non-salient weights, we perform binarization following Equation \eqref{bi-eq}. However, previous analytically derived scaling factors ignore implicit correlations among rows and directional shifts which cannot be accurately captured through mathematical derivation. To address this issue, we set scaling factors as learnable and propose a novel efficient block-wise optimization pipeline to learn them while considering implicit row-wise dependencies and angular biases, as demonstrated in the orange and green area of Figure \ref{PTQ1.61}.

In order to conduct an effective distance metric for optimization, we first consider MSE loss to reduce magnitude gaps. Then for angular biases described above, we take cosine similarity, a metric considering directional gaps, into account. We formulate the joint metric as follows:
\begin{equation}
\mathbb{E}(f_1,f_2) = \lVert f_1-f_2 \rVert_2 + \mathcal{D}_{NLC}(f_1,f_2),
\end{equation}
where $\mathbb{E}(\cdot)$ is the distance metric for optimization. $f_1$ and $f_2$ are different features. $D_{NLC}(\cdot)$ represents the negative logarithm of cosine similarity loss \citep{zhao2024lrquant}, as given by:
\begin{equation}
\mathcal{D}_{NLC}(f_1,f_2) = -\mbox{log}(\mathcal{C}(f_1,f_2)), 
\end{equation}
where $\mathcal{C}(\cdot)$ denotes cosine similarity. Followed by CBQ \citep{ding2023cbq}, our block-wise pipeline consists of two branches: the first branch aims at mitigating quantization error propagation and the second branch is tailored for quantifying the outputs distinction for the same inputs. Our final optimization objective is formulated as:
\begin{equation}
    \begin{aligned}
\operatorname*{argmin}_{{\alpha}_s^{*},{\alpha}_r^{*}}(\mathbb{E}(\mathcal{F}(\mbox{X},\mbox{W}),\mathcal{F}(\mbox{X}_{q},\mbox{W}_q^{'}))+ \\
\mathbb{E}(\mathcal{F}(\mbox{X}_q,\mbox{W}),\mathcal{F}(\mbox{X}_{q},\mbox{W}_q^{'}))),
\end{aligned}
\end{equation}
where ${\alpha}_s$ and ${\alpha}_r$ are scaling factors for magnitude and angular biases, respectively. $\mbox{W}_q^{'}$ denotes the quantized weights (see Appendix \ref{angular bias describe}) and $\mathcal{F}(\cdot)$ represents the embedding function of a block. $\mbox{X}$ indicates the input activation of the full-precision block while $\mbox{X}_q$ is that of the quantized block. 

With novel optimization strategy, \textbf{PTQ\textit{1.61}} outperforms previous low-bit PTQ methods significantly. The contribution is assessed in Table \ref{ablation} and Appendix \ref{appendix_c}.

\begin{table*}[t]
  \small
  \centering
  \setlength{\tabcolsep}{8pt}
  \begin{tabular}{@{\extracolsep{-4pt}}ccccccccccc}
    \toprule

     {\textbf{Dataset}} &{\textbf{Methods}} &{\textbf{Bits}} & 1-7  & 1-13 & 1-30 & 1-65 & 2-7 & 2-13 & 2-70 &3-8\\
    \midrule
    \multirow{8}{*}{WikiText2}&FP & 16 &5.68 &5.09 &4.10 &3.53 &5.47 &4.88 &3.31 &6.14\\
    \cmidrule{2-11}
    &AWQ & 2  &2.5e5 & 2.8e5 & 2.4e5 &7.4e4 & 2.2e5 &1.2e5 &- &8.2e5 \\
    &GPTQ     & 2 &2.1e3& 5.5e3 & 1.9e3 &55.91 & 7.7e3 &2.1e3 &77.95 &5.7e4 \\
    &QuIP & 2 &42.19 &12.18 &9.36 &7.19 &55.00 &13.75 & 6.96 & 119.23 \\
    &OmniQuant     & 2  &15.47& 13.21 & 8.81 & 7.58 & 37.37 &17.21 &7.81 & 796.8 4\\
    &PB-LLM     & 1.7(+1)  &102.19&48.11 &26.37 & 12.91 &66.30 &462.84 &NAN & 78.67 \\
    &BiLLM     & 1(+1.1) &35.04& 15.14 &10.52 & 8.51 &32.48 &21.77 &12.80 &50.59 \\
    &\cellcolor{gray!20}\textbf{PTQ\textit{1.61}}   & \cellcolor{gray!20}\textbf{1.61}  &\cellcolor{gray!20}\textbf{12.50} & \cellcolor{gray!20}\textbf{9.67} & \cellcolor{gray!20}\textbf{7.95} & \cellcolor{gray!20}\textbf{7.02} & \cellcolor{gray!20}\textbf{12.70} &\cellcolor{gray!20}\textbf{9.74} &\cellcolor{gray!20}\textbf{6.94} &\cellcolor{gray!20}\textbf{22.90} \\
    \cmidrule{1-11}
    \multirow{7}{*}{C4}&FP & 16 &7.08 &6.61 &5.98 &5.62 &6.97 &6.46 &5.52 &8.88\\
    \cmidrule{2-11}
    &AWQ & 2  &1.9e5 & 2.3e5 & 2.4e5 &7.5e4 & 1.7e5 &9.4e4 &- &8.1e5 \\
    &GPTQ     & 2  &689.13 & 2.5e3 & 234.95 &40.58 & NAN &323.12 &48.82 &1.0e5 \\
    &OmniQuant     & 2  &24.89& 18.31 &13.67 & 10.77 & 90.64 &26.76&12.28 &2.4e3 \\
    &PB-LLM     & 1.7(+1)  &67.92& 34.20 &22.45 & 13.70 & 66.23 &333.54 & NAN &78.98 \\
    &BiLLM     & 1(+1.1) &33.64& 14.75 &\textbf{10.97} & 9.92 &33.72 &23.14 &13.84 &48.26 \\
    &\cellcolor{gray!20}\textbf{PTQ\textit{1.61}}     &\cellcolor{gray!20}\textbf{1.61} &\cellcolor{gray!20}\textbf{17.13} & \cellcolor{gray!20}\textbf{13.51} & \cellcolor{gray!20}10.98 & \cellcolor{gray!20}\textbf{9.86} & \cellcolor{gray!20}\textbf{17.73} &\cellcolor{gray!20}\textbf{13.64} &\cellcolor{gray!20}\textbf{12.63} &\cellcolor{gray!20}\textbf{33.82} \\
    \bottomrule
  \end{tabular}
  \caption{Perplexities comparison of PTQ methods on LLaMA families. For PB-LLM and BiLLM, \textit{1.7(+1)} and \textit{1(+1.1)} under \textbf{Bits} means \textit{Weight bits(+Mask bits).} OPT results can be found in Table \ref{appendix-ppl}.}
  \label{ppl}
\end{table*}

\begin{figure}[t]
    \centering
    \includegraphics[width=3in]{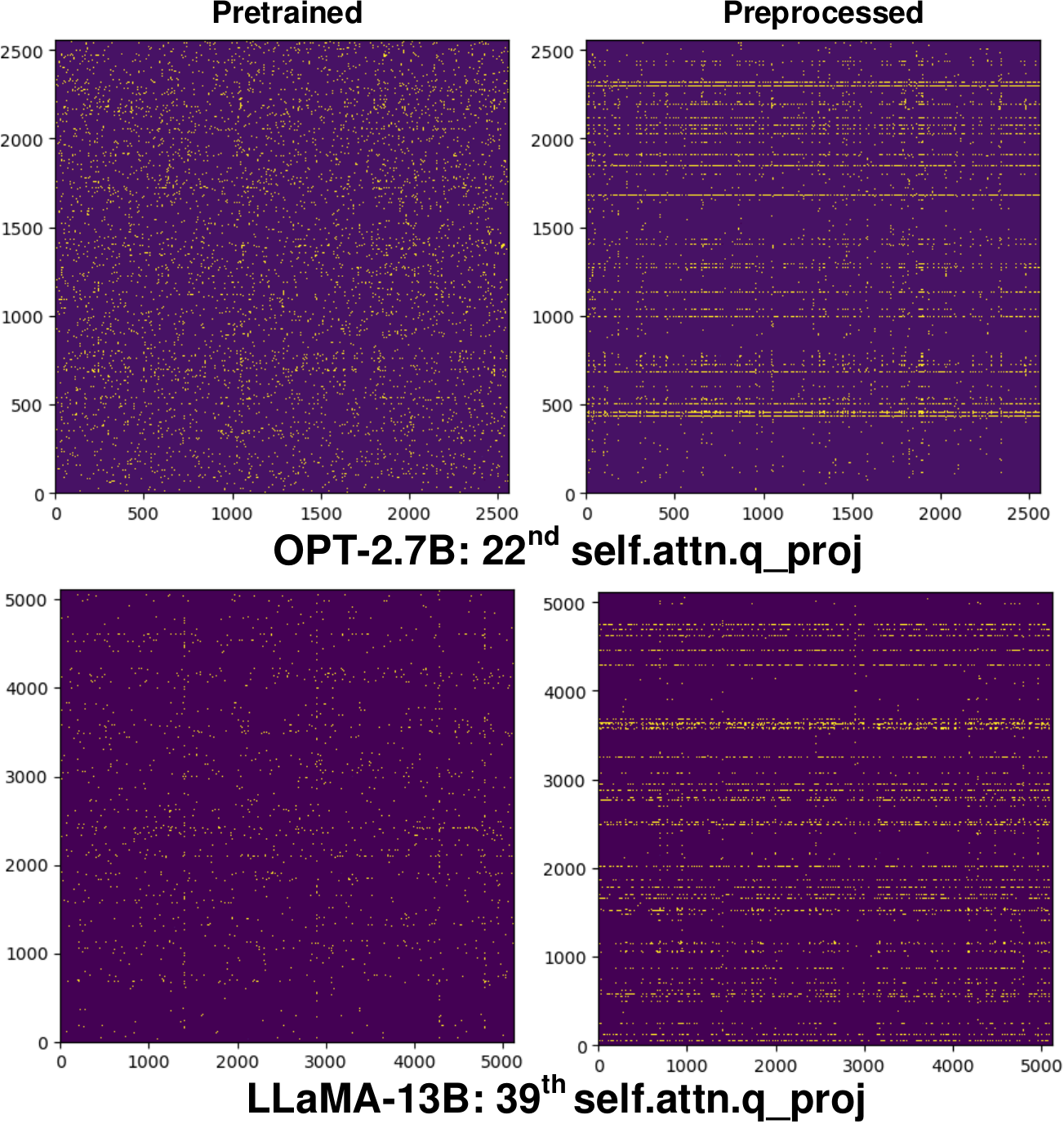}%
    \caption{Salient weights distribution of the pretrained and preprocessed OPT-2.7B and LLaMA-13B. The salient weights of the pretrained model exhibit a scattered distribution while after quantization preprocessing a visible row-wise concentrated pattern appears. Each row in the weight matrices corresponds to an input channel. More visualization results please refer to Figure \ref{more visual}.}
    \label{distribution}
\end{figure}

\subsection{Quantization Preprocessing}
\label{preprocessing}

It is well-known that the pretrained models exhibit the best performance so prior PTQ studies intuitively believe that quantizing the pretrained models should also yield optimal results. However, we discover that this notion is somewhat biased. As illustrated in Figure \ref{distribution}, we highlight the salient weights of a linear layer in LLMs based on magnitude-metric \citep{shang2023pb} where a scattered distribution pattern can be observed. The scattered pattern results in significant quantization errors when calculating row-wise scaling factors in per-channel quantization scheme. Therefore, we infer that under the premise of minimizing the bad impact on the pretrained model, transforming weights with similar saliency in a row-wise distribution pattern can effectively enhance the quantization performance.

As claimed by LoRA \citep{hu2021lora}, when fine-tuning a model on specific tasks, the weights compensation exhibits low-rank characteristics, indicating that fine-tuning may compensate for important information into specific dimensions of the weight matrices. Inspired by this proposition, we propose a novel quantization preprocessing paradigm. Specifically, we hope to perform a lightweight restorative LoRA on the pre-training dataset (\textit{i.e.}, RedPajama \citep{together2023redpajama}, the pre-training dataset of LLaMA families) to partially restore the performance of an initial quantized model to its pretrained version while transforming the distribution of salient weights into a concentrated row-wise pattern. As shown in Figure \ref{distribution}, it is evident that the salient weights in the preprocessed model indeed exhibit a concentrated row-wise pattern and the results in Table \ref{ablation} and \ref{appendix-ppl} demonstrate its effectiveness.


We emphasize that \textbf{the goal of our preprocessing scheme is to produce a LLM that is more amenable to channel-wise quantization, rather than using the preprocessed model for inference directly}. In Appendix \ref{pre vs post}, we further illustrate this point with example experiments. Additionally, it is crucial to clarify that our preprocessing scheme stands out significant advantages and differences from existing post-quantization PEFT methods, e.g., LoRA, QLoRA and QA-LoRA \citep{xu2023qa}, please refer to Appendix \ref{appendix_d}.

\section{Experiments}
In this section, we conduct extensive experiments to validate our novel extremely low-bit PTQ method \textbf{PTQ\textit{1.61}} on various benchmarks and LLMs with existing methods to demonstrate that our approach achieves outstanding performance under extremely challenging quantization. 

\subsection{Experimental Setup}

\paragraph{Baseline}

Since our \textbf{PTQ\textit{1.61}} is an extremely low-bit PTQ method, we primarily choose PB-LLM (10\% 8-bit) \citep{shang2023pb} and BiLLM \citep{huang2024billm}, which claim to be extremely low-bit methods but actually have an equivalent bit-width larger than 2-bit, as baselines. Additionally, several state-of-the-art PTQ methods (2-bit) such as OmniQuant \citep{OmniQuant}, AWQ \citep{lin2023awq}, QuIP \citep{chee2024quip}, and GPTQ \citep{frantar2022gptq} are also be evaluated. 

\paragraph{Models}
We evaluate our method mainly on LLaMA \citep{touvron2023llama1}, LLaMA-2 \citep{touvron2023llama2} and LLaMA-3 \citep{dubey2024llama}, for LLaMA-families are currently the most popular and widely applied among LLMs. Considering the comprehensiveness, experiments on OPT families \citep{zhang2022opt} are in Appendix \ref{appendix_d}.

\begin{table*}[t]
  \small
  \centering
  \setlength{\tabcolsep}{5pt}
  \begin{tabular}{cccccccccccc}
    \toprule
    \textbf{LLaMA}& \textbf{Methods} &\textbf{Bits} &PIQA & ARC-e & HellaS & Wing & Race & ARC-c & LAMB-o &LAMB-s & \textbf{Avg.}\\
    \midrule
    \multirow{7}{*}{1-8} &FP &16 &78.67 &75.29 &56.99 &70.01 &40.29 &41.81 &73.57 &67.82 & 63.06\\
    \cmidrule{2-12}
    &GPTQ &2 &53.64 &26.09 &25.87 &47.75&22.68&22.44&0.0&0.0&24.81\\
     &QuIP &2 &60.23 &38.26 &34.83 &51.22 &28.90 &22.10 &15.10 &8.52 &32.40\\
     &OmniQuant &2 &58.22&39.94&32.45 &52.49 &32.25 &22.10 &23.66 &12.77 &34.24\\
     &PB-LLM &1.7(+1) &55.71&29.12&28.31&48.86 &26.41 &19.80 &10.42 &10.09 &28.59 \\
     &BiLLM &1(+1.1) &61.10&40.99&31.80 &53.67 &30.14 &20.64 &23.15 &16.48 &36.00\\ 
     &\cellcolor{gray!20}\textbf{PTQ\textit{1.61}} &\cellcolor{gray!20}\textbf{1.61} &\cellcolor{gray!20} \textbf{63.71}&\cellcolor{gray!20} \textbf{49.62}&\cellcolor{gray!20}\textbf{35.73} &\cellcolor{gray!20}\textbf{56.75}  &\cellcolor{gray!20}\textbf{32.54}  &\cellcolor{gray!20}\textbf{25.26} &\cellcolor{gray!20}\textbf{38.93} &\cellcolor{gray!20}\textbf{26.57} &\cellcolor{gray!20}\textbf{41.14}\\
    \midrule
    \multirow{6}{*}{1-13} &FP & 16 &79.16 &77.36 &59.92 &72.77 &39.62 & 46.42 &76.15 &71.08 &65.31 \\
    \cmidrule{2-12}
    &GPTQ &2 &51.90&25.84&26.08 &49.72 &24.11 & 22.18 &0.0 &0.0 &24.98\\
     &OmniQuant &2 &67.14&51.43&\textbf{41.28} &56.20 &32.73 & 29.52 &23.40 &17.85 &39.94\\
     &PB-LLM &1.7(+1) &60.45&37.46&30.79 &51.07 &30.24 &18.69 &27.71 &21.33 &34.72\\
     &BiLLM &1(+1.1) &67.90 &50.84&39.02 &\textbf{62.19} & 34.07 &26.11 &\textbf{49.93} &33.75 &44.98\\
     &\cellcolor{gray!20}\textbf{PTQ\textit{1.61}} &\cellcolor{gray!20}\textbf{1.61} &\cellcolor{gray!20}\textbf{68.17} &\cellcolor{gray!20}\textbf{58.59} &\cellcolor{gray!20}40.02 &\cellcolor{gray!20}58.33 &\cellcolor{gray!20}\textbf{34.26} &\cellcolor{gray!20}\textbf{27.22} &\cellcolor{gray!20}45.95 &\cellcolor{gray!20}\textbf{35.94} &\cellcolor{gray!20}\textbf{46.56}\\
     \midrule
    \multirow{6}{*}{1-30} &FP &16 &80.96 &80.39 &63.34 &75.69 &40.57 &52.82 &77.59 &73.34 &68.09\\
    \cmidrule{2-12}
    &GPTQ &2 &52.50&20.39&25.88 &51.38 &23.25 &21.33 &0.06 &0.0 &24.35\\
     &QuIP &2 &\textbf{72.42} &58.80 &45.64 &63.30&35.69 &30.20 &52.45 &36.64 &49.39\\
     &OmniQuant &2 &70.35&58.03&44.82 &58.17 &34.93 &31.74 &41.88 &31.75 &46.46\\
     &PB-LLM &1.7(+1) &63.76&40.11&33.32 &61.17 &30.91 &21.33 &43.78 &33.09 &40.93\\
     &\cellcolor{gray!20}\textbf{PTQ\textit{1.61}} &\cellcolor{gray!20}\textbf{1.61} &\cellcolor{gray!20}70.24&\cellcolor{gray!20}\textbf{63.64}&\cellcolor{gray!20}\textbf{46.82} &\cellcolor{gray!20}\textbf{63.61} &\cellcolor{gray!20}\textbf{37.13} &\cellcolor{gray!20}\textbf{32.17} &\cellcolor{gray!20}\textbf{55.95} &\cellcolor{gray!20}\textbf{44.61} &\cellcolor{gray!20}\textbf{51.77}\\
    \midrule
    \multirow{6}{*}{2-7}
    & FP & 16 & 78.07 &76.30 & 57.14 &69.06 &39.52 &43.34 &73.86 &68.23 & 63.19\\
    \cmidrule{2-12}
     &QuIP &2 &56.53 &28.70 &27.52 &48.78 &24.40 &18.94 &3.18 &2.06 &26.26\\
     &OmniQuant &2 &57.34&38.80&30.11&51.78&27.37&20.73&3.98&1.47 &30.20\\
     &PB-LLM &1.7(+1) &54.46&28.20&27.03&49.09&26.70&19.11&7.08&5.71&27.17\\
     &BiLLM &1(+1.1) &60.39&39.94&30.74&51.93&29.57&21.16&18.44&14.05&33.28\\
     &\cellcolor{gray!20}\textbf{PTQ\textit{1.61}} &\cellcolor{gray!20}\textbf{1.61} &\cellcolor{gray!20}\textbf{63.22} &\cellcolor{gray!20}\textbf{47.18} &\cellcolor{gray!20}\textbf{35.78} &\cellcolor{gray!20}\textbf{52.25} &\cellcolor{gray!20}\textbf{29.86} &\cellcolor{gray!20}\textbf{22.27} &\cellcolor{gray!20}\textbf{37.38} &\cellcolor{gray!20}\textbf{25.65} &\cellcolor{gray!20}\textbf{39.20}\\
    \midrule
    \multirow{6}{*}{2-13}
    & FP &16 &79.11 &79.46 &60.04 &72.14 &40.57 &48.46 &76.77 &70.33 &65.86 \\
    \cmidrule{2-12}
     &QuIP &2 &65.45 &51.56 &39.65 &55.72 &31.58 &25.85 &33.86 &22.67 &40.79\\
     &OmniQuant &2 &62.62&44.27 &40.16 &52.17 &30.81 &24.66 &20.07 &10.17 &35.62\\
     &PB-LLM &1.7(+1) &54.46&27.95&26.74&49.96&26.03&19.54&3.14&2.50&26.29\\
     &BiLLM &1(+1.1) &63.55 &49.83 &34.36 &\textbf{58.17} &32.34 &23.81 &40.81 &25.15 &41.00\\
     &\cellcolor{gray!20}\textbf{PTQ\textit{1.61}} &\cellcolor{gray!20}\textbf{1.61} &\cellcolor{gray!20}\textbf{66.54} &\cellcolor{gray!20}\textbf{56.86} &\cellcolor{gray!20}\textbf{40.32} &\cellcolor{gray!20}55.88 &\cellcolor{gray!20}\textbf{33.30} &\cellcolor{gray!20}\textbf{26.45} &\cellcolor{gray!20}\textbf{47.23} &\cellcolor{gray!20}\textbf{31.21} &\cellcolor{gray!20}\textbf{44.72}\\
    \midrule
    \multirow{7}{*}{3-8} 
    &FP & 16 &79.54 &80.09 &60.14 &73.24 &40.29 &50.17 &75.65 &68.72 &65.98\\
    \cmidrule{2-12}
    &GPTQ &2 &52.39 &26.14 &25.74 &51.54 &20.19 &20.65 &0.0 &0.0 &24.58 \\
     &QuIP &2 &52.72 &26.43 &29.32 &50.67 &27.75 &20.39 &4.93 &3.01 &26.90\\
     &OmniQuant &2 &54.13 &28.87 &26.50 &50.12 &22.68 &20.48 &0.02 &0.02 &25.35 \\
     &PB-LLM &1.7(+1) &56.91&32.37&28.43&49.25&27.66&17.41&16.63&12.44&30.14  \\
     &BiLLM &1(+1.1) &60.01&38.26&31.48&\textbf{53.75}&\textbf{30.14}&19.45&26.30&15.51&34.36 \\
     &\cellcolor{gray!20}\textbf{PTQ\textit{1.61}} &\cellcolor{gray!20}\textbf{1.61} &\cellcolor{gray!20}\textbf{63.22} &\cellcolor{gray!20}\textbf{46.17}& \cellcolor{gray!20}\textbf{34.71} &\cellcolor{gray!20}52.80 &\cellcolor{gray!20}29.09 &\cellcolor{gray!20}\textbf{23.04} &\cellcolor{gray!20}\textbf{30.27} &\cellcolor{gray!20}\textbf{18.26}&\cellcolor{gray!20}\textbf{37.20}\\
    \bottomrule
  \end{tabular}
    \caption{Reasoning accuracies comparison on LLaMA family. More tasks are listed in Table \ref{mmlu/gsm8k} and \ref{longbench}.}
  \label{zeroshot1}
\end{table*}

\paragraph{Training Details}
We initialize learnable scaling factors with $\alpha_w = \frac{\lVert w \rVert_1}{n_{w}}$ and AdamW optimizer \citep{loshchilov2017decoupled} with zero weight decay is utilized to update them with learning rate 5e-4 and 1e-3. For PTQ, our calibration set sampled from WikiText2 \citep{merity2016pointer} consists of 128 random 2048 token-segments and the block-wise training process includes 20 epochs with a batch size of 1. For quantization preprocessing, the number of steps and ranks in lightweight restorative LoRA is 20K and 64 respectively. The entire process is deployed on 2 Nvidia A800 GPUs.

\paragraph{Datasets}
On language generation tasks which are the core objectives of LLMs, our test set comes from WikiText2 and C4 \citep{raffel2020exploring}. We also select several reasoning benchmarks, \textit{i.e.}, PIQA \citep{bisk2020piqa}, ARC \citep{clark2018think}, HellaSwag \citep{zellers2019hellaswag}, Winogrande \citep{sakaguchi2021winogrande}, Race \citep{lai-etal-2017-race} and LAMBADA \citep{paperno2016lambada}, using the open-sourced toolkit lm-evaluation-harness \citep{eval-harness}.  We also assess the evaluation on MMLU \citep{hendryckstest2021}, GSM8K \citep{cobbe2021training} and LongBench \citep{bai2024longbench}, please refer to Appendix \ref{more-eval}. In addition, RedPajama \citep{together2023redpajama} is used for quantization preprocessing.

\begin{figure*}[t]
  \centering
  \includegraphics[width=5in]{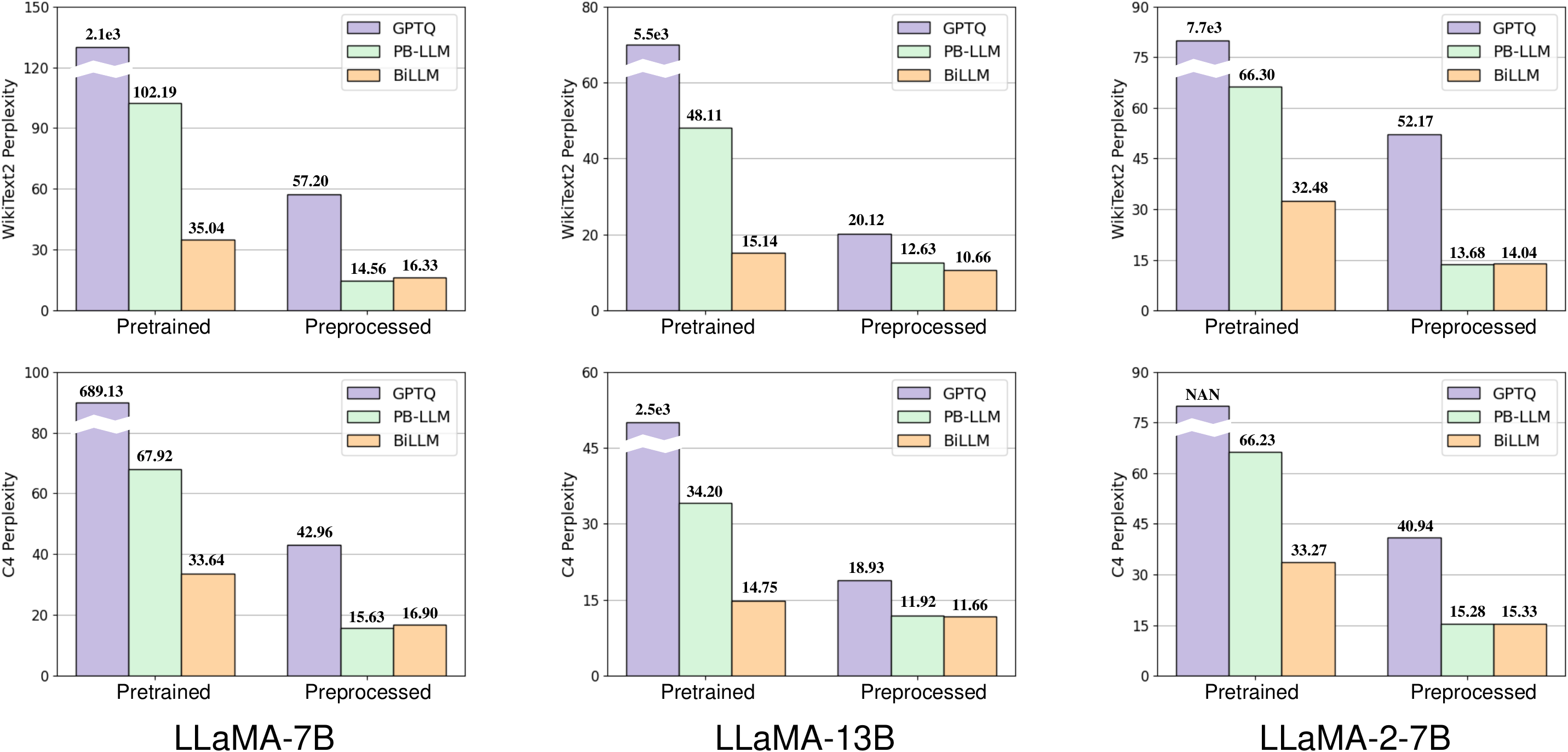}%
  \caption{Our novel quantization preprocessing scheme on other existing PTQ methods. Results on more LLMs and common sense reasoning tasks can be found in Figure \ref{bar-chart-opt}.}
  \label{bar-chart}
\end{figure*}

\subsection{Experiments on Language Generation Tasks}
The fundamental prowess of LLMs lies in their language generation capabilities. Consequently, evaluating such capabilities of a quantized model via perplexity serves as the core metric of a quantization method. As presented in Table \ref{ppl}, we compare the perplexities between our \textbf{PTQ\textit{1.61}} and other baselines to valid the effectiveness on extremely low-bit PTQ tasks, from where we can see that our method achieves promising performance. In comparison to the two extremely low-bit methods PB-LLM and BiLLM, our performance significantly surpasses theirs while not introducing intolerable bit-width for each weight. In terms of OmniQuant, which performs best on LLaMA families among baselines, our method still surpasses it by a significant margin. For instance, on LLaMA-2-7B we achieves $12.70$ while OmniQuant is $37.37$ in WikiText2. Significantly, our method performs better on LLaMA3, which is known to be harder to quantize.

\subsection{Experiments on Reasoning Benchmarks}

Reasoning capability is becoming a crucial metric for evaluating PTQ approaches. The comparison results are indicated in Table \ref{zeroshot1}, where our method exhibits superiority in most benchmarks. For instance, compared with the second best baseline, BiLLM, our method showcases an average performance increase of $1.58\% \sim 5.92\%$. Particularly, on LAMBADA most baselines suffer from significant performance degradation while we still maintain an outstanding level. Considering the comparison results and the minimal 1.61-bit, we conclude that our method represents the state-of-the-art extremely low-bit PTQ scheme for LLMs.


\subsection{Ablation Study}

After demonstrating the advancement of our \textbf{PTQ\textit{1.61}}, we conduct ablation study on LLaMA-13B to further validate the effectiveness of our each innovation as indicated in Table \ref{ablation}. As the first row, without any additional improvements, directly using the derived analytically scaling factors for binarization would almost entirely compromise the text generation capability of LLMs. When utilizing our structured masks to retain salient weights as the second row, there is a significant improvement, indicating the importance of salient weights and the effectiveness of our masks, but there remains considerable room for enhancement. Furthermore, our novel block-wise strategy for non-salient weights binarization lifts the performance to a excellent level as demonstrated by the forth row. Ultimately, the last row illustrates that the row-wise pattern obtained by our quantization preprocessing brings a remarkable enhancement. More detailed ablation results are available in Appendix \ref{appendix_b}.

\begin{table}[t]
\small
\centering
\setlength{\tabcolsep}{2.2pt}
\begin{tabular}{ccccc}
    \toprule
\textbf{Structured} & \textbf{Learnable} & \multirow{2}{*}{\textbf{Preprocess}}  & \multirow{2}{*}{WikiText2} & \multirow{2}{*}{C4} \\
 \textbf{Mask}  &\textbf{Scalar} & & \\
\midrule
\textbf{-}&\textbf{-} &\textbf{-} & 14664 &11377\\
\checkmark &\textbf{-} &\textbf{-} & 1370.4&772.83\\
\textbf{-} &\textbf{-} &\checkmark & 569.81&702.44\\
\checkmark &\checkmark &\textbf{-} & 14.22&20.78\\
\checkmark &\checkmark & \checkmark &\textbf{9.67}&\textbf{13.51}\\
    \bottomrule
\end{tabular}
\caption{\label{ablation}
Ablation study (PPL) on LLaMA-13B.}
\end{table}

\subsection{Quantization Preprocessing on Baselines}
\label{preprocess-baseline}
In addition to our \textbf{PTQ\textit{1.61}}, we also employ the proposed quantization preprocessing scheme on other baselines to validate its scalability and the results can be found in Figure \ref{bar-chart}, which demonstrate that significant improvements appear in all baselines, especially for 2-bit GPTQ which completely collapses without preprocessing. With the effectiveness of our preprocessing scheme, future research can take a fresh perspective to focus on finding a more appropriately pretrained model. 

At present, despite its strong performance, our preprocessing scheme still suffers from drawbacks such as longer runtime (as indicated in Table \ref{cost}). Therefore, we must clarify that our preprocessing scheme is viewed as an \textbf{optional component} of our \textbf{PTQ\textit{1.61}}. As indicated in Table \ref{appendix-ppl}, even without preprocessing, our \textbf{PTQ\textit{1.61}} still achieves state-of-the-art performance under extremely low-bit setting.

\section{Conclusion}
In this paper, we explore the real limit of post-training quantization and propose an extremely low-bit PTQ approach namely \textbf{PTQ\textit{1.61}}, which is truly the first PTQ method enables sub 2-bit quantization for LLMs. Firstly, one-dimensional structured mask with negligibly additional 0.0002-bit per weight is introduced to preserve salient weights. For non-salient weights binarization, we devise an efficient block-wise optimization strategy to learn scaling factors considering row correlations and angular biases. In addition to above contributions, we further propose a quantization preprocessing paradigm to transform the salient weights into a row-wise pattern which is able to alleviate the difficulty in per-channel quantization. Extensive experiments indicate that \textbf{PTQ\textit{1.61}} becomes state-of-the-art extremely low-bit PTQ method for LLMs.

\section*{Limitation}

Although showcasing superior performance, the preprocessing scheme still has limitations to be reckoned with, which requires more runtime to get a start point before quantization. For example, our runtime reaches 2h on LLaMA-7B, and fortunately, this falls within an acceptable range (OmniQuant reports 1.1h but exhibits worse performance and higher bit-width per weight). Considering that extremely low-bit quantization is the most challenging quantization scenario especially for PTQ, we believe it is worthwhile to sacrifice some computational resources within an acceptable range to pursue higher performance. 

In addition, due to the limitation that commercial NVIDIA GPUs do not support such low-bit inference, and designing specific hardware requires larger research teams and financial support, we cannot provide real-world inference evaluation results yet. Our goal is to explore the performance limits of PTQ by fake-quantization before commercial hardware support is available. We believe this will eventually be realized as evidenced by the quick development of GPUs.

\section*{Ethics Statement}

This paper introduces solutions to the challenges associated with Large Language Models (LLMs) quantization, with the overarching goal of facilitating the widespread adoption and application of LLMs. In the current landscape, ethical concerns tied to LLMs, including the presence of hidden biases encoded in the models, are garnering heightened attention. Following our investigation, we assert that our proposed method does not further amplify the biases and contravene any ethical standards.

\section*{Acknowledgment}
Miao Zhang was partially sponsored by the National Natural Science Foundation of China under Grant 62306084 and U23B2051, and Shenzhen College Stability Support Plan under Grant GXWD20231128102243003 and Grant ZDSYS20230626091203008.

\bibliography{acl_latex}

\begin{thebibliography}{62}
\expandafter\ifx\csname natexlab\endcsname\relax\def\natexlab#1{#1}\fi

\bibitem[{Bai et~al.(2024)Bai, Lv, Zhang, Lyu, Tang, Huang, Du, Liu, Zeng, Hou, Dong, Tang, and Li}]{bai2024longbench}
Yushi Bai, Xin Lv, Jiajie Zhang, Hongchang Lyu, Jiankai Tang, Zhidian Huang, Zhengxiao Du, Xiao Liu, Aohan Zeng, Lei Hou, Yuxiao Dong, Jie Tang, and Juanzi Li. 2024.
\newblock \href {https://doi.org/10.18653/v1/2024.acl-long.172} {{L}ong{B}ench: A bilingual, multitask benchmark for long context understanding}.
\newblock In \emph{Proceedings of the 62nd Annual Meeting of the Association for Computational Linguistics (Volume 1: Long Papers)}, pages 3119--3137, Bangkok, Thailand. Association for Computational Linguistics.

\bibitem[{Bisk et~al.(2020)Bisk, Zellers, Gao, Choi et~al.}]{bisk2020piqa}
Yonatan Bisk, Rowan Zellers, Jianfeng Gao, Yejin Choi, et~al. 2020.
\newblock Piqa: Reasoning about physical commonsense in natural language.
\newblock In \emph{Proceedings of the AAAI conference on artificial intelligence}, volume~34, pages 7432--7439.

\bibitem[{Brown et~al.(2020)Brown, Mann, Ryder, Subbiah, Kaplan, Dhariwal, Neelakantan, Shyam, Sastry, Askell et~al.}]{brown2020language}
Tom Brown, Benjamin Mann, Nick Ryder, Melanie Subbiah, Jared~D Kaplan, Prafulla Dhariwal, Arvind Neelakantan, Pranav Shyam, Girish Sastry, Amanda Askell, et~al. 2020.
\newblock Language models are few-shot learners.
\newblock \emph{Advances in neural information processing systems}, 33:1877--1901.

\bibitem[{Bulat and Tzimiropoulos(2019)}]{bulat2019xnor}
Adrian Bulat and Georgios Tzimiropoulos. 2019.
\newblock Xnor-net++: Improved binary neural networks.
\newblock \emph{arXiv preprint arXiv:1909.13863}.

\bibitem[{Chan and Ioannidis(1998)}]{chan1998bitmap}
Chee-Yong Chan and Yannis~E Ioannidis. 1998.
\newblock Bitmap index design and evaluation.
\newblock In \emph{Proceedings of the 1998 ACM SIGMOD international conference on Management of data}, pages 355--366.

\bibitem[{Chee et~al.(2024)Chee, Cai, Kuleshov, and De~Sa}]{chee2024quip}
Jerry Chee, Yaohui Cai, Volodymyr Kuleshov, and Christopher~M De~Sa. 2024.
\newblock Quip: 2-bit quantization of large language models with guarantees.
\newblock \emph{Advances in Neural Information Processing Systems}, 36.

\bibitem[{Clark et~al.(2019)Clark, Khandelwal, Levy, and Manning}]{clark2019does}
Kevin Clark, Urvashi Khandelwal, Omer Levy, and Christopher~D Manning. 2019.
\newblock What does bert look at? an analysis of bert's attention.
\newblock \emph{arXiv preprint arXiv:1906.04341}.

\bibitem[{Clark et~al.(2018)Clark, Cowhey, Etzioni, Khot, Sabharwal, Schoenick, and Tafjord}]{clark2018think}
Peter Clark, Isaac Cowhey, Oren Etzioni, Tushar Khot, Ashish Sabharwal, Carissa Schoenick, and Oyvind Tafjord. 2018.
\newblock Think you have solved question answering? try arc, the ai2 reasoning challenge.
\newblock \emph{arXiv preprint arXiv:1803.05457}.

\bibitem[{Cobbe et~al.(2021)Cobbe, Kosaraju, Bavarian, Chen, Jun, Kaiser, Plappert, Tworek, Hilton, Nakano et~al.}]{cobbe2021training}
Karl Cobbe, Vineet Kosaraju, Mohammad Bavarian, Mark Chen, Heewoo Jun, Lukasz Kaiser, Matthias Plappert, Jerry Tworek, Jacob Hilton, Reiichiro Nakano, et~al. 2021.
\newblock Training verifiers to solve math word problems.
\newblock \emph{arXiv preprint arXiv:2110.14168}.

\bibitem[{Computer(2023)}]{together2023redpajama}
Together Computer. 2023.
\newblock \href {https://github.com/togethercomputer/RedPajama-Data} {Redpajama: an open dataset for training large language models}.

\bibitem[{Courbariaux et~al.(2016)Courbariaux, Hubara, Soudry, El-Yaniv, and Bengio}]{courbariaux2016binarized}
Matthieu Courbariaux, Itay Hubara, Daniel Soudry, Ran El-Yaniv, and Yoshua Bengio. 2016.
\newblock Binarized neural networks: Training deep neural networks with weights and activations constrained to+ 1 or-1.
\newblock \emph{arXiv preprint arXiv:1602.02830}.

\bibitem[{Dettmers et~al.(2023)Dettmers, Pagnoni, Holtzman, and Zettlemoyer}]{dettmers2023qlora}
Tim Dettmers, Artidoro Pagnoni, Ari Holtzman, and Luke Zettlemoyer. 2023.
\newblock Qlora: Efficient finetuning of quantized llms.
\newblock \emph{arXiv preprint arXiv:2305.14314}.

\bibitem[{Ding et~al.(2023)Ding, Liu, Zhang, Tu, Li, Hu, Chen, Tang, Xiong, Yin et~al.}]{ding2023cbq}
Xin Ding, Xiaoyu Liu, Yun Zhang, Zhijun Tu, Wei Li, Jie Hu, Hanting Chen, Yehui Tang, Zhiwei Xiong, Baoqun Yin, et~al. 2023.
\newblock Cbq: Cross-block quantization for large language models.
\newblock \emph{arXiv preprint arXiv:2312.07950}.

\bibitem[{Dubey et~al.(2024)Dubey, Jauhri, Pandey, Kadian, Al-Dahle, Letman, Mathur, Schelten, Yang, Fan et~al.}]{dubey2024llama}
Abhimanyu Dubey, Abhinav Jauhri, Abhinav Pandey, Abhishek Kadian, Ahmad Al-Dahle, Aiesha Letman, Akhil Mathur, Alan Schelten, Amy Yang, Angela Fan, et~al. 2024.
\newblock The llama 3 herd of models.
\newblock \emph{arXiv preprint arXiv:2407.21783}.

\bibitem[{Esser et~al.(2019)Esser, McKinstry, Bablani, Appuswamy, and Modha}]{esser2019learned}
Steven~K Esser, Jeffrey~L McKinstry, Deepika Bablani, Rathinakumar Appuswamy, and Dharmendra~S Modha. 2019.
\newblock Learned step size quantization.
\newblock \emph{arXiv preprint arXiv:1902.08153}.

\bibitem[{Frantar and Alistarh(2023)}]{frantar2023sparsegpt}
Elias Frantar and Dan Alistarh. 2023.
\newblock Sparsegpt: Massive language models can be accurately pruned in one-shot.
\newblock In \emph{International Conference on Machine Learning}, pages 10323--10337. PMLR.

\bibitem[{Frantar et~al.(2022)Frantar, Ashkboos, Hoefler, and Alistarh}]{frantar2022gptq}
Elias Frantar, Saleh Ashkboos, Torsten Hoefler, and Dan Alistarh. 2022.
\newblock Gptq: Accurate post-training quantization for generative pre-trained transformers.
\newblock \emph{arXiv preprint arXiv:2210.17323}.

\bibitem[{Gao et~al.(2023)Gao, Tow, Abbasi, Biderman, Black, DiPofi, Foster, Golding, Hsu, Le~Noac'h, Li, McDonell, Muennighoff, Ociepa, Phang, Reynolds, Schoelkopf, Skowron, Sutawika, Tang, Thite, Wang, Wang, and Zou}]{eval-harness}
Leo Gao, Jonathan Tow, Baber Abbasi, Stella Biderman, Sid Black, Anthony DiPofi, Charles Foster, Laurence Golding, Jeffrey Hsu, Alain Le~Noac'h, Haonan Li, Kyle McDonell, Niklas Muennighoff, Chris Ociepa, Jason Phang, Laria Reynolds, Hailey Schoelkopf, Aviya Skowron, Lintang Sutawika, Eric Tang, Anish Thite, Ben Wang, Kevin Wang, and Andy Zou. 2023.
\newblock \href {https://doi.org/10.5281/zenodo.10256836} {A framework for few-shot language model evaluation}.

\bibitem[{Gou et~al.(2021)Gou, Yu, Maybank, and Tao}]{gou2021knowledge}
Jianping Gou, Baosheng Yu, Stephen~J Maybank, and Dacheng Tao. 2021.
\newblock Knowledge distillation: A survey.
\newblock \emph{International Journal of Computer Vision}, 129(6):1789--1819.

\bibitem[{Hendrycks et~al.(2021)Hendrycks, Burns, Basart, Zou, Mazeika, Song, and Steinhardt}]{hendryckstest2021}
Dan Hendrycks, Collin Burns, Steven Basart, Andy Zou, Mantas Mazeika, Dawn Song, and Jacob Steinhardt. 2021.
\newblock Measuring massive multitask language understanding.
\newblock \emph{Proceedings of the International Conference on Learning Representations (ICLR)}.

\bibitem[{Hu et~al.(2021)Hu, Shen, Wallis, Allen-Zhu, Li, Wang, Wang, and Chen}]{hu2021lora}
Edward~J Hu, Yelong Shen, Phillip Wallis, Zeyuan Allen-Zhu, Yuanzhi Li, Shean Wang, Lu~Wang, and Weizhu Chen. 2021.
\newblock Lora: Low-rank adaptation of large language models.
\newblock \emph{arXiv preprint arXiv:2106.09685}.

\bibitem[{Huang et~al.(2019)Huang, Ni, and Yang}]{huang2019efficient}
Kun Huang, Bingbing Ni, and Xiaokang Yang. 2019.
\newblock Efficient quantization for neural networks with binary weights and low bitwidth activations.
\newblock In \emph{Proceedings of the AAAI Conference on Artificial Intelligence}, volume~33, pages 3854--3861.

\bibitem[{Huang et~al.(2024)Huang, Liu, Qin, Li, Zhang, Liu, Magno, and Qi}]{huang2024billm}
Wei Huang, Yangdong Liu, Haotong Qin, Ying Li, Shiming Zhang, Xianglong Liu, Michele Magno, and Xiaojuan Qi. 2024.
\newblock Billm: Pushing the limit of post-training quantization for llms.
\newblock \emph{arXiv preprint arXiv:2402.04291}.

\bibitem[{Lai et~al.(2017)Lai, Xie, Liu, Yang, and Hovy}]{lai-etal-2017-race}
Guokun Lai, Qizhe Xie, Hanxiao Liu, Yiming Yang, and Eduard Hovy. 2017.
\newblock \href {https://doi.org/10.18653/v1/D17-1082} {{RACE}: Large-scale {R}e{A}ding comprehension dataset from examinations}.
\newblock In \emph{Proceedings of the 2017 Conference on Empirical Methods in Natural Language Processing}, pages 785--794, Copenhagen, Denmark. Association for Computational Linguistics.

\bibitem[{Lee et~al.(2024)Lee, Jin, Kim, Kim, and Park}]{lee2024owq}
Changhun Lee, Jungyu Jin, Taesu Kim, Hyungjun Kim, and Eunhyeok Park. 2024.
\newblock Owq: Outlier-aware weight quantization for efficient fine-tuning and inference of large language models.
\newblock In \emph{Proceedings of the AAAI Conference on Artificial Intelligence}, volume~38, pages 13355--13364.

\bibitem[{Li et~al.(2021)Li, Gong, Tan, Yang, Hu, Zhang, Yu, Wang, and Gu}]{li2021brecq}
Yuhang Li, Ruihao Gong, Xu~Tan, Yang Yang, Peng Hu, Qi~Zhang, Fengwei Yu, Wei Wang, and Shi Gu. 2021.
\newblock Brecq: Pushing the limit of post-training quantization by block reconstruction.
\newblock \emph{arXiv preprint arXiv:2102.05426}.

\bibitem[{Lin et~al.(2023)Lin, Tang, Tang, Yang, Dang, and Han}]{lin2023awq}
Ji~Lin, Jiaming Tang, Haotian Tang, Shang Yang, Xingyu Dang, and Song Han. 2023.
\newblock Awq: Activation-aware weight quantization for llm compression and acceleration.
\newblock \emph{arXiv preprint arXiv:2306.00978}.

\bibitem[{Lin et~al.(2020)Lin, Ji, Xu, Zhang, Wang, Wu, Huang, and Lin}]{lin2020rotated}
Mingbao Lin, Rongrong Ji, Zihan Xu, Baochang Zhang, Yan Wang, Yongjian Wu, Feiyue Huang, and Chia-Wen Lin. 2020.
\newblock Rotated binary neural network.
\newblock \emph{Advances in neural information processing systems}, 33:7474--7485.

\bibitem[{Liu et~al.(2022)Liu, Niu, Yuan, Yang, Wang, and Liu}]{liu2022pd}
Jiawei Liu, Lin Niu, Zhihang Yuan, Dawei Yang, Xinggang Wang, and Wenyu Liu. 2022.
\newblock Pd-quant: Post-training quantization based on prediction difference metric.
\newblock \emph{arXiv preprint arXiv:2212.07048}.

\bibitem[{Liu et~al.(2023{\natexlab{a}})Liu, Liu, Gao, Gao, Zhao, Li, Ding, and Wen}]{liu2023emergent}
Peiyu Liu, Zikang Liu, Ze-Feng Gao, Dawei Gao, Wayne~Xin Zhao, Yaliang Li, Bolin Ding, and Ji-Rong Wen. 2023{\natexlab{a}}.
\newblock Do emergent abilities exist in quantized large language models: An empirical study.
\newblock \emph{arXiv preprint arXiv:2307.08072}.

\bibitem[{Liu et~al.(2023{\natexlab{b}})Liu, Oguz, Zhao, Chang, Stock, Mehdad, Shi, Krishnamoorthi, and Chandra}]{liu2023llm}
Zechun Liu, Barlas Oguz, Changsheng Zhao, Ernie Chang, Pierre Stock, Yashar Mehdad, Yangyang Shi, Raghuraman Krishnamoorthi, and Vikas Chandra. 2023{\natexlab{b}}.
\newblock Llm-qat: Data-free quantization aware training for large language models.
\newblock \emph{arXiv preprint arXiv:2305.17888}.

\bibitem[{Liu et~al.(2018)Liu, Wu, Luo, Yang, Liu, and Cheng}]{liu2018bi}
Zechun Liu, Baoyuan Wu, Wenhan Luo, Xin Yang, Wei Liu, and Kwang-Ting Cheng. 2018.
\newblock Bi-real net: Enhancing the performance of 1-bit cnns with improved representational capability and advanced training algorithm.
\newblock In \emph{Proceedings of the European conference on computer vision (ECCV)}, pages 722--737.

\bibitem[{Longpre et~al.(2023)Longpre, Hou, Vu, Webson, Chung, Tay, Zhou, Le, Zoph, Wei et~al.}]{longpre2023flan}
Shayne Longpre, Le~Hou, Tu~Vu, Albert Webson, Hyung~Won Chung, Yi~Tay, Denny Zhou, Quoc~V Le, Barret Zoph, Jason Wei, et~al. 2023.
\newblock The flan collection: Designing data and methods for effective instruction tuning.
\newblock \emph{arXiv preprint arXiv:2301.13688}.

\bibitem[{Loshchilov and Hutter(2017)}]{loshchilov2017decoupled}
Ilya Loshchilov and Frank Hutter. 2017.
\newblock Decoupled weight decay regularization.
\newblock \emph{arXiv preprint arXiv:1711.05101}.

\bibitem[{Ma et~al.(2024)Ma, Wang, Ma, Wang, Wang, Huang, Dong, Wang, Xue, and Wei}]{ma2024era}
Shuming Ma, Hongyu Wang, Lingxiao Ma, Lei Wang, Wenhui Wang, Shaohan Huang, Li~Dong, Ruiping Wang, Jilong Xue, and Furu Wei. 2024.
\newblock The era of 1-bit llms: All large language models are in 1.58 bits.
\newblock \emph{arXiv preprint arXiv:2402.17764}.

\bibitem[{Ma et~al.(2023)Ma, Fang, and Wang}]{ma2023llm}
Xinyin Ma, Gongfan Fang, and Xinchao Wang. 2023.
\newblock Llm-pruner: On the structural pruning of large language models.
\newblock \emph{Advances in neural information processing systems}, 36:21702--21720.

\bibitem[{Merity et~al.(2016)Merity, Xiong, Bradbury, and Socher}]{merity2016pointer}
Stephen Merity, Caiming Xiong, James Bradbury, and Richard Socher. 2016.
\newblock Pointer sentinel mixture models.
\newblock \emph{arXiv preprint arXiv:1609.07843}.

\bibitem[{Nagel et~al.(2020)Nagel, Amjad, Van~Baalen, Louizos, and Blankevoort}]{nagel2020up}
Markus Nagel, Rana~Ali Amjad, Mart Van~Baalen, Christos Louizos, and Tijmen Blankevoort. 2020.
\newblock Up or down? adaptive rounding for post-training quantization.
\newblock In \emph{International Conference on Machine Learning}, pages 7197--7206. PMLR.

\bibitem[{Ouyang et~al.(2022)Ouyang, Wu, Jiang, Almeida, Wainwright, Mishkin, Zhang, Agarwal, Slama, Ray et~al.}]{ouyang2022training}
Long Ouyang, Jeffrey Wu, Xu~Jiang, Diogo Almeida, Carroll Wainwright, Pamela Mishkin, Chong Zhang, Sandhini Agarwal, Katarina Slama, Alex Ray, et~al. 2022.
\newblock Training language models to follow instructions with human feedback.
\newblock \emph{Advances in neural information processing systems}, 35:27730--27744.

\bibitem[{Paperno et~al.(2016)Paperno, Kruszewski, Lazaridou, Pham, Bernardi, Pezzelle, Baroni, Boleda, and Fern{\'a}ndez}]{paperno2016lambada}
Denis Paperno, Germ{\'a}n Kruszewski, Angeliki Lazaridou, Quan~Ngoc Pham, Raffaella Bernardi, Sandro Pezzelle, Marco Baroni, Gemma Boleda, and Raquel Fern{\'a}ndez. 2016.
\newblock The lambada dataset: Word prediction requiring a broad discourse context.
\newblock \emph{arXiv preprint arXiv:1606.06031}.

\bibitem[{Qin et~al.(2020)Qin, Gong, Liu, Bai, Song, and Sebe}]{qin2020binary}
Haotong Qin, Ruihao Gong, Xianglong Liu, Xiao Bai, Jingkuan Song, and Nicu Sebe. 2020.
\newblock Binary neural networks: A survey.
\newblock \emph{Pattern Recognition}, 105:107281.

\bibitem[{Raffel et~al.(2020)Raffel, Shazeer, Roberts, Lee, Narang, Matena, Zhou, Li, and Liu}]{raffel2020exploring}
Colin Raffel, Noam Shazeer, Adam Roberts, Katherine Lee, Sharan Narang, Michael Matena, Yanqi Zhou, Wei Li, and Peter~J Liu. 2020.
\newblock Exploring the limits of transfer learning with a unified text-to-text transformer.
\newblock \emph{Journal of machine learning research}, 21(140):1--67.

\bibitem[{Rastegari et~al.(2016)Rastegari, Ordonez, Redmon, and Farhadi}]{rastegari2016xnor}
Mohammad Rastegari, Vicente Ordonez, Joseph Redmon, and Ali Farhadi. 2016.
\newblock Xnor-net: Imagenet classification using binary convolutional neural networks.
\newblock In \emph{European conference on computer vision}, pages 525--542. Springer.

\bibitem[{Sakaguchi et~al.(2021)Sakaguchi, Bras, Bhagavatula, and Choi}]{sakaguchi2021winogrande}
Keisuke Sakaguchi, Ronan~Le Bras, Chandra Bhagavatula, and Yejin Choi. 2021.
\newblock Winogrande: An adversarial winograd schema challenge at scale.
\newblock \emph{Communications of the ACM}, 64(9):99--106.

\bibitem[{Shang et~al.(2023)Shang, Yuan, Wu, and Dong}]{shang2023pb}
Yuzhang Shang, Zhihang Yuan, Qiang Wu, and Zhen Dong. 2023.
\newblock Pb-llm: Partially binarized large language models.
\newblock \emph{arXiv preprint arXiv:2310.00034}.

\bibitem[{Shao et~al.(2023)Shao, Chen, Zhang, Xu, Zhao, Li, Zhang, Gao, Qiao, and Luo}]{OmniQuant}
Wenqi Shao, Mengzhao Chen, Zhaoyang Zhang, Peng Xu, Lirui Zhao, Zhiqian Li, Kaipeng Zhang, Peng Gao, Yu~Qiao, and Ping Luo. 2023.
\newblock Omniquant: Omnidirectionally calibrated quantization for large language models.
\newblock \emph{arXiv preprint arXiv:2308.13137}.

\bibitem[{Taori et~al.(2023)Taori, Gulrajani, Zhang, Dubois, Li, Guestrin, Liang, and Hashimoto}]{alpaca}
Rohan Taori, Ishaan Gulrajani, Tianyi Zhang, Yann Dubois, Xuechen Li, Carlos Guestrin, Percy Liang, and Tatsunori~B. Hashimoto. 2023.
\newblock Stanford alpaca: An instruction-following llama model.
\newblock \url{https://github.com/tatsu-lab/stanford_alpaca}.

\bibitem[{Touvron et~al.(2023{\natexlab{a}})Touvron, Lavril, Izacard, Martinet, Lachaux, Lacroix, Rozi{\`e}re, Goyal, Hambro, Azhar et~al.}]{touvron2023llama1}
Hugo Touvron, Thibaut Lavril, Gautier Izacard, Xavier Martinet, Marie-Anne Lachaux, Timoth{\'e}e Lacroix, Baptiste Rozi{\`e}re, Naman Goyal, Eric Hambro, Faisal Azhar, et~al. 2023{\natexlab{a}}.
\newblock Llama: Open and efficient foundation language models.
\newblock \emph{arXiv preprint arXiv:2302.13971}.

\bibitem[{Touvron et~al.(2023{\natexlab{b}})Touvron, Martin, Stone, Albert, Almahairi, Babaei, Bashlykov, Batra, Bhargava, Bhosale et~al.}]{touvron2023llama2}
Hugo Touvron, Louis Martin, Kevin Stone, Peter Albert, Amjad Almahairi, Yasmine Babaei, Nikolay Bashlykov, Soumya Batra, Prajjwal Bhargava, Shruti Bhosale, et~al. 2023{\natexlab{b}}.
\newblock Llama 2: Open foundation and fine-tuned chat models.
\newblock \emph{arXiv preprint arXiv:2307.09288}.

\bibitem[{Tunstall et~al.(2023)Tunstall, Beeching, Lambert, Rajani, Rasul, Belkada, Huang, von Werra, Fourrier, Habib et~al.}]{tunstall2023zephyr}
Lewis Tunstall, Edward Beeching, Nathan Lambert, Nazneen Rajani, Kashif Rasul, Younes Belkada, Shengyi Huang, Leandro von Werra, Cl{\'e}mentine Fourrier, Nathan Habib, et~al. 2023.
\newblock Zephyr: Direct distillation of lm alignment.
\newblock \emph{arXiv preprint arXiv:2310.16944}.

\bibitem[{Vig and Belinkov(2019)}]{vig2019analyzing}
Jesse Vig and Yonatan Belinkov. 2019.
\newblock Analyzing the structure of attention in a transformer language model.
\newblock \emph{arXiv preprint arXiv:1906.04284}.

\bibitem[{Wang et~al.(2023)Wang, Ma, Dong, Huang, Wang, Ma, Yang, Wang, Wu, and Wei}]{wang2023bitnet}
Hongyu Wang, Shuming Ma, Li~Dong, Shaohan Huang, Huaijie Wang, Lingxiao Ma, Fan Yang, Ruiping Wang, Yi~Wu, and Furu Wei. 2023.
\newblock Bitnet: Scaling 1-bit transformers for large language models.
\newblock \emph{arXiv preprint arXiv:2310.11453}.

\bibitem[{Wei et~al.(2022)Wei, Zhang, Zhang, Gong, Zhang, Zhang, Yu, and Liu}]{wei2022outlier}
Xiuying Wei, Yunchen Zhang, Xiangguo Zhang, Ruihao Gong, Shanghang Zhang, Qi~Zhang, Fengwei Yu, and Xianglong Liu. 2022.
\newblock Outlier suppression: Pushing the limit of low-bit transformer language models.
\newblock \emph{Advances in Neural Information Processing Systems}, 35:17402--17414.

\bibitem[{Xiao et~al.(2023)Xiao, Lin, Seznec, Wu, Demouth, and Han}]{xiao2023smoothquant}
Guangxuan Xiao, Ji~Lin, Mickael Seznec, Hao Wu, Julien Demouth, and Song Han. 2023.
\newblock Smoothquant: Accurate and efficient post-training quantization for large language models.
\newblock In \emph{International Conference on Machine Learning}, pages 38087--38099. PMLR.

\bibitem[{Xu et~al.(2023)Xu, Xie, Gu, Chen, Chang, Zhang, Chen, Zhang, and Tian}]{xu2023qa}
Yuhui Xu, Lingxi Xie, Xiaotao Gu, Xin Chen, Heng Chang, Hengheng Zhang, Zhensu Chen, Xiaopeng Zhang, and Qi~Tian. 2023.
\newblock Qa-lora: Quantization-aware low-rank adaptation of large language models.
\newblock \emph{arXiv preprint arXiv:2309.14717}.

\bibitem[{Xu et~al.(2024)Xu, Han, Yang, Wang, Zhu, Liu, Liu, and Che}]{xu2024onebit}
Yuzhuang Xu, Xu~Han, Zonghan Yang, Shuo Wang, Qingfu Zhu, Zhiyuan Liu, Weidong Liu, and Wanxiang Che. 2024.
\newblock Onebit: Towards extremely low-bit large language models.
\newblock \emph{arXiv preprint arXiv:2402.11295}.

\bibitem[{Xu et~al.(2021)Xu, Lin, Liu, Chen, Shao, Gao, Tian, and Ji}]{xu2021recu}
Zihan Xu, Mingbao Lin, Jianzhuang Liu, Jie Chen, Ling Shao, Yue Gao, Yonghong Tian, and Rongrong Ji. 2021.
\newblock Recu: Reviving the dead weights in binary neural networks.
\newblock In \emph{Proceedings of the IEEE/CVF international conference on computer vision}, pages 5198--5208.

\bibitem[{Yao et~al.(2022)Yao, Yazdani~Aminabadi, Zhang, Wu, Li, and He}]{yao2022zeroquant}
Zhewei Yao, Reza Yazdani~Aminabadi, Minjia Zhang, Xiaoxia Wu, Conglong Li, and Yuxiong He. 2022.
\newblock Zeroquant: Efficient and affordable post-training quantization for large-scale transformers.
\newblock \emph{Advances in Neural Information Processing Systems}, 35:27168--27183.

\bibitem[{Yuan et~al.(2023)Yuan, Niu, Liu, Liu, Wang, Shang, Sun, Wu, Wu, and Wu}]{yuan2023rptq}
Zhihang Yuan, Lin Niu, Jiawei Liu, Wenyu Liu, Xinggang Wang, Yuzhang Shang, Guangyu Sun, Qiang Wu, Jiaxiang Wu, and Bingzhe Wu. 2023.
\newblock Rptq: Reorder-based post-training quantization for large language models.
\newblock \emph{arXiv preprint arXiv:2304.01089}.

\bibitem[{Zellers et~al.(2019)Zellers, Holtzman, Bisk, Farhadi, and Choi}]{zellers2019hellaswag}
Rowan Zellers, Ari Holtzman, Yonatan Bisk, Ali Farhadi, and Yejin Choi. 2019.
\newblock Hellaswag: Can a machine really finish your sentence?
\newblock \emph{arXiv preprint arXiv:1905.07830}.

\bibitem[{Zhang et~al.(2022)Zhang, Roller, Goyal, Artetxe, Chen, Chen, Dewan, Diab, Li, Lin et~al.}]{zhang2022opt}
Susan Zhang, Stephen Roller, Naman Goyal, Mikel Artetxe, Moya Chen, Shuohui Chen, Christopher Dewan, Mona Diab, Xian Li, Xi~Victoria Lin, et~al. 2022.
\newblock Opt: Open pre-trained transformer language models.
\newblock \emph{arXiv preprint arXiv:2205.01068}.

\bibitem[{Zhao et~al.(2024)Zhao, Zhang, Zeng, Wang, Liu, and Nie}]{zhao2024lrquant}
Jiaqi Zhao, Miao Zhang, Chao Zeng, Ming Wang, Xuebo Liu, and Liqiang Nie. 2024.
\newblock Lrquant: Learnable and robust post-training quantization for large language models.
\newblock In \emph{Proceedings of the 62nd Annual Meeting of the Association for Computational Linguistics (Volume 1: Long Papers)}, pages 2240--2255.

\end{thebibliography}

\clearpage

\appendix

\section*{Appendix}
\label{sec:appendix}

\section{Average Bit-width Per Weight of Linear Layer}
\label{appendix_a}

For a weight in a mix-precision quantized linear layer, its average bit-width $b$ is calculated by following formulation:
\begin{equation}
b=1*r_b + b_{salient} * (1-r_b) + b_{index} + b_{additional},
\label{bitwidth}
\end{equation}
where $r_b$ is the ratio of binarized weights in the layer and $b_{salient}$ denotes the bit-width of salient weights. The first two item is also called weight bit-width. $b_{index}$ represents the bit-width for index storing using the bitmap mechanism \citep{chan1998bitmap} and $b_{additional}$ is used for saving quantization parameters, \textit{i.e.}, scaling factors. 

Assume the weight matrix is $4096 \times 4096$ in such layer. For our \textbf{PTQ\textit{1.61}} which saves $20\%$ salient weights to 4-bit and binaries the others, the weight bit-width can be effortlessly determined as 1.6-bit and the overall bits number is $4096\times4096\times0.8+4096\times4096\times0.8\times4=26,843,545$. In addition, the shape of our 1-bit one-dimensional structured mask is $4096\times1$, so its $r_{b}$ is $1\times4096\div26,843,545\approx0.0002$. Moreover, quantization parameters in our method contains 3 low-dimensional scaling factors and a part of zero-points, so $b_{additional}$ will be $(3\times4096\times1\times16+0.2\times4096\times16)\div26,843,545\approx0.008$. Overall, the average bit-width per weight in a layer quantized by our \textbf{PTQ\textit{1.61}} is $b=1.6+0.0002+0.008\approx1.61$.

For PB-LLM which selects $10\%$ salient weights at 8-bit using a 1-bit fine-grained unstructured mask with the same shape as the weight matrix, the obtained average bit-width per weight is $b=0.1\times8+0.9\times1+1=2.7$.

BiLLM devises a finer-grained binarization scheme which divides all weights into $3$ groups and calculates group-wise scaling factors. Specifically, they propose a structured mask based on Hessian for salient weights and an unstructured mask based magnitude for unsalient weights. From their paper we get that their weight bit-width is 1-bit and $b_{additonal}$ is 0.1-bit then we have $b=1.0+0.1+1.0=2.1$.

\section{Structured Mask}
\label{appendix_b}
\subsection{The Impact of Salient Channels Ratio in Proposed Structured Mask}

\begin{figure}[t]
    \centering
    \includegraphics[width=0.4\textwidth]{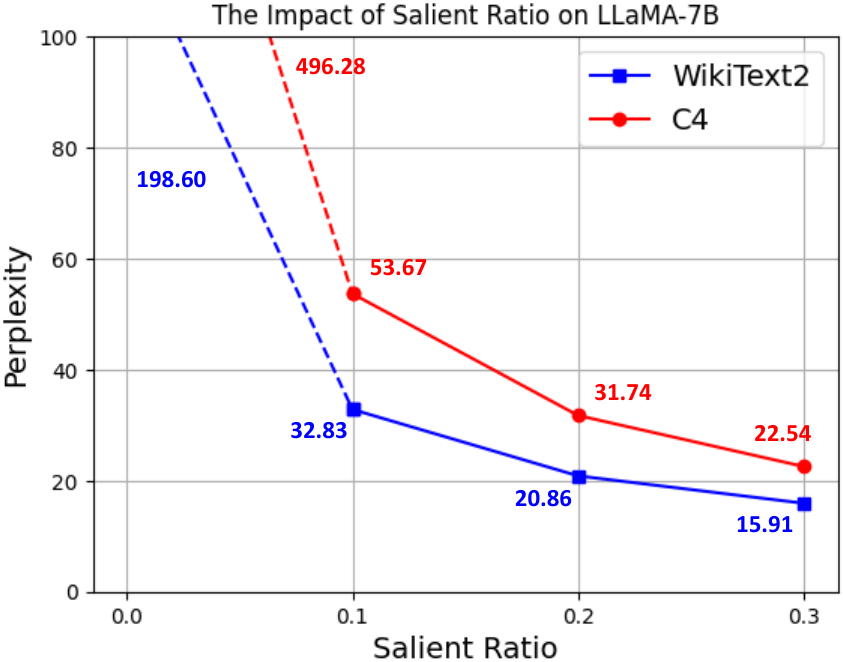}
    \caption{The impact of salient ratio in \textbf{PTQ\textit{1.61}} on LLaMA-7B.}
    \label{line-chart}
\end{figure}

Inspired by PB-LLM which declares the importance of salient weights, we devise a one-dimensional structured mask to preserve top $20\%$ salient channels of weight matrices at 4-bit based on input activations in our \textbf{PTQ\textit{1.61}}. More comprehensively, we delve into the effects of the salient channels ratio on the quantized model. As illustrated in Figure \ref{line-chart}, we compare the performance of \textbf{PTQ\textit{1.61}} on pretrained LLaMA-7B with different ratios and the results indicate that higher salient ratios lead to better performance. The reason why we give up the optimal $30\%$ is that the average bit-width per weight in the quantized model nearly approaches 2-bit (1.91-bit), which violates the conditions for extremely low-bit quantization. Therefore, we opt for the second-best performance $20\%$ as our preserved ratio. However, it is crucial to note that the aforementioned experiments do not diminish the significance of our other innovations on non-salient weights binarization. As evident from the Table \ref{ablation}, simply maintaining the salient weights without incorporating block-wise optimization will results in inadequate performance.

\subsection{Distinctions with AWQ and OWQ}
For AWQ, both their method and our \textbf{PTQ\textit{1.61}} take into account the relationship between input activation and weight, and utilize this relationship for subsequent processing. But notably, there is no structured mask in AWQ. They leverage this discovery to perform a grid search based on MSE loss to select appropriate quantization scalars. These scalars are used to scale the corresponding weights up or down. Weights corresponding to larger input activations are assigned larger scalars to increase their magnitude to reduce quantization error. Conversely, our structured mask is designed through rigorous mathematical derivation that input activation has a greater impact on the upper bound of quantization error. This allows us to assess the importance of corresponding weights and avoid unacceptable binarization errors. Furthermore, the results in Table 5 demonstrate that our method outperforms AWQ in extremely low-bit quantization scenarios without requiring additional innovations.

\begin{table}[t]
\small
  \centering
  \begin{tabular}{ccccc}
    \toprule
    \textbf{LLaMA}& \textbf{Methods} &\textbf{Bits} &WikiText2 & C4\\
    \midrule
    \multirow{2}{*}{1-7}&OWQ &2 &13.64&\textbf{15.78}\\
     &\textbf{PTQ\textit{1.61}} &\textbf{1.61} &\textbf{12.50}&17.13\\
    \midrule
    \multirow{2}{*}{1-13}&OWQ &2 &10.69&\textbf{12.45}\\
     &\textbf{PTQ\textit{1.61}} &\textbf{1.61} &\textbf{9.67}&13.51\\
    \midrule
    \multirow{2}{*}{2-7}&OWQ &2 &18.78&19.93\\
     &\textbf{PTQ\textit{1.61}} &\textbf{1.61} &\textbf{12.70}&\textbf{17.73}\\
    \midrule
    \multirow{2}{*}{2-13}&OWQ &2 &30.28&36.78\\
     &\textbf{PTQ\textit{1.61}} &\textbf{1.61} &\textbf{9.74}&\textbf{13.64}\\
    \bottomrule
  \end{tabular}
    \caption{PPL comparison between OWQ and \textbf{PTQ\textit{1.61}} on LLaMA family.}
  \label{owq1}
\end{table}

\begin{table}[t]
\small
  \centering
  \begin{tabular}{ccccc}
    \toprule
    \textbf{LLaMA}& \textbf{Methods} &\textbf{Mask} &WikiText2 & C4\\
    \midrule
    \multirow{2}{*}{1-7}&\multirow{2}{*}{\textbf{PTQ\textit{1.61}}} &OWQ &22.11&33.77\\
     & &Ours &\textbf{12.50}&\textbf{17.13}\\
    \midrule
    \multirow{2}{*}{1-13}&\multirow{2}{*}{\textbf{PTQ\textit{1.61}}} &OWQ &57.33&132.08\\
     &  &Ours &\textbf{9.67}&\textbf{13.51}\\
    \midrule
    \multirow{2}{*}{2-7}&\multirow{2}{*}{\textbf{PTQ\textit{1.61}}} &OWQ &NAN&NAN\\
     &  &Ours &\textbf{12.70}&\textbf{17.73}\\
    \midrule
    \multirow{2}{*}{2-13}&\multirow{2}{*}{\textbf{PTQ\textit{1.61}}} &OWQ &1.1e3&2.7e3\\
     &  &Ours &\textbf{9.74}&\textbf{13.64}\\
    \bottomrule
  \end{tabular}
    \caption{PPL comparison between OWQ and our mask.}
  \label{owq2}
\end{table}

For OWQ, we need to clarify that the two method differ significantly in terms of innovations, motivation, and the tasks they target. In the process of formula derivation, OWQ use Hessian matrix (related to input activation) to decompose quantization errors to determine which weights columns to retain at full precision. However, it is important to emphasize that there are approximation processes in decomposing, such as ignoring the first-order and higher-order terms in Taylor expansion. Moreover, using Cholesky decomposition to simulate the Gaussian elimination of Hessian updating also involves approximations. In contrast, our structured mask does not involve approximation in its mathematical derivation. It directly derives two key factors that affect the upper bound of quantization errors and combined with visualization, determines that input activation has the greatest impact on the upper bound of quantization errors. This avoids introducing additional errors due to mathematical calculations. Therefore, while there are similarities in form between the structured mask of our \textbf{PTQ\textit{1.61}} and OWQ, the motivations are entirely different. Approximation in the Hessian-based structured mask is reasonable for high-bit quantization. However, in extremely low-bit PTQ, especially binarization, these approximations can be infinitely magnified. Thus, directly reducing the upper bound of quantization errors is a method with fewer errors. To prove this, we set up comparative experiments on several models as Table \ref{owq1}, where our \textbf{PTQ\textit{1.61}} quantizes the model to 1.61-bit, while OWQ quantizes it to 2-bit. The results show that the models quantized by PTQ1.61 achieve better results under lower bit-width, especially on the LLaMA2. We also design structured masks based on Hessian as OWQ to preserve 20\% salient channels, from Table \ref{owq2} it can be observed that compared with our \textbf{PTQ\textit{1.61}} the performance of structured mask proposed by OWQ collapses, which also proves our structured mask has smaller errors and is more suitable for extremely low-bit quantization. Furthermore, it is noteworthy that OWQ needs to store certain weights in FP16 format. Due to the differences in storage methods between INT and FP formats, designing kernel functions is very difficult. In contrast, our weights are all in INT format, making kernel function design much simpler.

\section{Block-wise Optimization}
\label{appendix_c}
\subsection{Hyperparameters}

In addition to the preservation ratio of the structured mask, the remaining hyperparameters are only the learning rates. We conduct extensive experiments with various learning rates from 1e-4 to 1e-2 and finally select the optimal. The limited number of hyperparameters also demonstrate that our \textbf{PTQ\textit{1.61}} is not complex and easy to deploy.

\begin{table*}[t]
\small

  \centering
  \setlength{\tabcolsep}{8pt}
  \begin{tabular}{@{\extracolsep{-4pt}}ccccccccccc}
    \toprule
    \multirow{2}{*}{\textbf{Dataset}}& \multirow{2}{*}{\textbf{Methods}} & \multirow{2}{*}{\textbf{Bits}} & \multicolumn{5}{c}{\textbf{LLaMA}} & \multicolumn{3}{c}{\textbf{OPT}} \\
    \cmidrule(r){4-8} \cmidrule(l){9-11}
     & & & 1-7  & 1-13 & 1-30 & 2-7 & 2-13 & 2.7 &6.7 &13 \\
    \midrule
    \multirow{6}{*}{WikiText2}&FP & 16 &5.68 &5.09 &4.10 &5.47 &4.88 &12.47 &10.86 &10.12\\
    \cmidrule{2-11}
    &OmniQuant     & 2  &15.47& 13.21 & 8.81 & 37.37 &17.21 &1.1e6&9.3e5&4.6e4 \\
    &PB-LLM     & 1.7(+1)  &102.19&48.11 &26.37 &66.30 &462.84 &238.18&174.76&75.28 \\
    &BiLLM     & 1(+1.1) &35.04& 15.14 &9.96 &32.48 &21.77 &49.55&45.36&18.22 \\
    &\textbf{PTQ\textit{1.61}*}  & \textbf{1.61}  & 20.86 &14.22& 11.84 & 22.58 & 15.63 &44.10 &27.51&22.94 \\
    &\textbf{PTQ\textit{1.61}}     & \textbf{1.61}  &\textbf{12.50} & \textbf{9.67} & \textbf{7.95} & \textbf{12.70} &\textbf{9.74} &\textbf{28.56} &\textbf{19.45} &\textbf{15.55}\\
    \cmidrule{1-11}
    \multirow{6}{*}{C4}&FP & 16 &7.08 &6.61 &5.98 &6.97 &6.46 &13.16 &11.74 &11.19 \\
    \cmidrule{2-11}
    &OmniQuant     & 2  &24.89& 18.31 &13.67 & 90.64 &26.76&9.4e5 &4.4e6&1.2e5 \\
    &PB-LLM     & 1.7(+1)  &67.92& 34.20 &22.45 & 66.23 &333.54 &161.47&102.85&47.50 \\
    &BiLLM     & 1(+1.1) &33.64& 14.75 &\textbf{10.95} &33.72 &23.14 &40.57&39.58&\textbf{17.78} \\
    &\textbf{PTQ\textit{1.61}*}     & \textbf{1.61}  &31.74&20.78 &15.53 & 36.07 &22.77 &72.64&38.37&29.97 \\
    &\textbf{PTQ\textit{1.61}}     &\textbf{1.61} &\textbf{17.13} & \textbf{13.51} & 10.98 & \textbf{17.73} &\textbf{13.64} &\textbf{33.45} &\textbf{22.78} &18.31 \\
    \bottomrule
  \end{tabular}
    \caption{Perplexities comparison of our \textbf{PTQ\textit{1.61}}, BiLLM and OmniQuant on pretrained and preprocessed LLaMA and OPT families. \textbf{*} indicates \textbf{PTQ\textit{1.61}} without preprocessed.}
    \label{appendix-ppl}
\end{table*}

\subsection{Angular Biases}
\label{angular bias describe}
Except for implict row-wise dependencies, our block-wise strategy also takes angular biases into account. Previous PTQ methods (OmniQuant, GPTQ or other quantization methods for CNN models such as BNN) only focuses on the magnitude gaps between FP models and their quantized counterpart, while the inherent directional distinctions, which has been proved to exist by RBNN and LRQuant, cannot be ignored which will not be addressed by traditional scaling factors and MSE loss. Therefore, we introduce loss function considering cosine similarity into block-wise optimization. We provide the detailed formula of dequantized weights considering scaling factors:
\begin{equation}
\mbox{W}_{q}^{'} = (\alpha_{r_1} \times \alpha_{r_2}) \circ (\alpha_{s}* \text{sign}(\mbox{W})).
\label{scaling angula}
\end{equation}
We conduct a comparison on whether this loss function is used or not in our \textbf{PTQ\textit{1.61}} in Table \ref{angular} and the results demonstrate that with this consideration, the block-wise optimization strategy used in our method becomes more robust and advanced than that in CBQ or OmniQuant. 

\begin{table}[t]
\small
  \centering
  \setlength{\tabcolsep}{12pt}
  \begin{tabular}{cccc}
    \toprule
    \textbf{LLaMA}&\textbf{Mask} &WikiText2 & C4\\
    \midrule
    \multirow{2}{*}{1-7B} &w/o &13.56&18.13\\
     &w &\textbf{12.50}&\textbf{17.13}\\
    \midrule
    \multirow{2}{*}{1-13B}&w/o &9.98&13.85\\
     &w &\textbf{9.67}&\textbf{13.51}\\
    \midrule
    \multirow{2}{*}{2-7B}&w/o &13.69&19.85\\
     &w &\textbf{12.70}&\textbf{17.73}\\
    \midrule
    \multirow{2}{*}{2-13B} &w/o &10.24&14.26\\
     &w &\textbf{9.74}&\textbf{13.64}\\
    \bottomrule
  \end{tabular}
    \caption{PPL comparison between whether considering angular biases (w) or not (w/o).}
  \label{angular}
\end{table}

\section{More Details on Quantization Preprocessing}
\label{appendix_d}

\subsection{Enhancement on Our PTQ\textit{1.61}}
In addition to LLaMA-13B in Table \ref{ablation}, we evaluate the enhancements on our \textbf{PTQ\textit{1.61}} brought by novel quantization preprocessing paradigm on more LLMs and the results are listed in Table \ref{appendix-ppl}, from which we confirm that the preprocessing consistently enhances the performance of our \textbf{PTQ\textit{1.61}} on each model. Crucially, our results reveal that, apart from preprocessing, our other innovations alone offer comparable performance advantages over existing methods, while attaining a lower weight compression ratio. 

\begin{figure*}[t]
  \centering
  \includegraphics[width=5.4in]{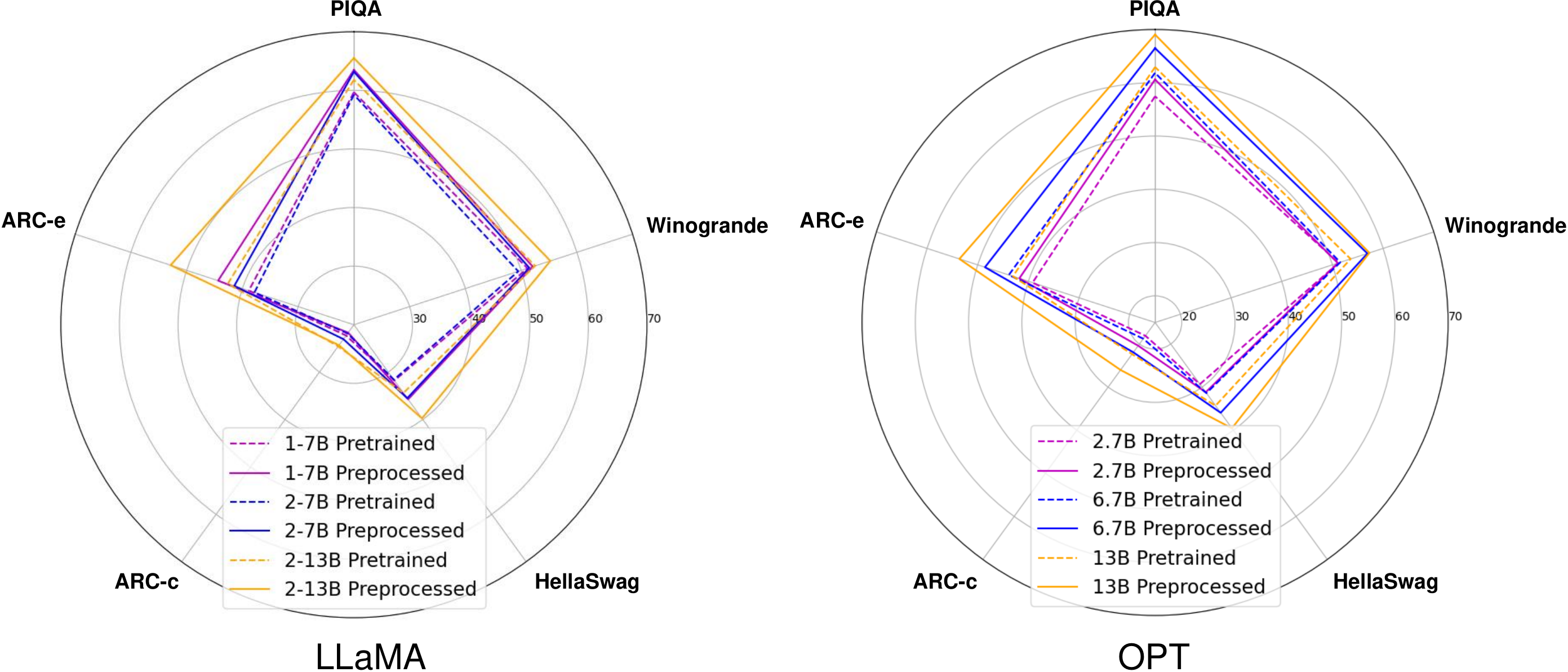}%
  \caption{Zero-shot accuracies comparison between pretrained and preprocessed model quantized by our \textbf{PTQ\textit{1.61}}.}
  \label{radar}
\end{figure*}

Furthermore, the quantization preprocessing also augments the common sense reasoning capabilites of our method as listed in Figure \ref{radar}.

\begin{table}[t]
  \small
  \centering
  \setlength{\tabcolsep}{3pt}
  \begin{tabular}{cccc}
    \toprule
    \textbf{LLaMA}&\textbf{Method} &\textbf{GPU Memory} & \textbf{Runtime}\\
    \midrule
    \multirow{3}{*}{7B} &OmniQ &13GB&1.1h\\
     &OneBit &360GB&24days\\
     &\textbf{PTQ\textit{1.61}} &15GB&2h\\
    \midrule
    \multirow{3}{*}{13B}&OmniQ &18GB&2.2h\\
     &OneBit &360GB&32days\\
     &\textbf{PTQ\textit{1.61}} &19GB&4.2h\\
    \bottomrule
  \end{tabular}
    \caption{Resource requirements comparison.}
  \label{cost}
\end{table}

\subsection{Preprocessed Model and FP Model}
\label{pre vs post}
In Section \ref{preprocessing}, we introduce the rationale behind our quantization preprocessing scheme and emphasize its purpose is to reshape the weight distribution to be more suitable for quantization while not using the preprocessed model directly for inference. As shown in Table \ref{f!=fp}, it can be observed that after preprocessing the FP16 performance of LLaMA-13B degrades slightly, but after quantization it significantly outperforms the pretrained model, which further proves the effectiveness of our preprocessing method.

\begin{table}
\small
    \caption{Full-precision perplexities comparison of the pretrained and preprocessed LLaMA-13B.}
    \label{f!=fp}
	\centering
    \begin{tabular}{cccc}
    \toprule
    \multirow{2}{*}{Model} &\multirow{2}{*}{Bits} & \multicolumn{2}{c}{Dataset}       \\
    \cmidrule(l){3-4}
      & & WikiText2   & C4 \\
    \midrule
    Pretrained &16 & 5.09  & 6.61     \\
    -Quant &1.61 & 14.22& 20.78\\
    \midrule
    Preprocessed &16 & 9.32 & 12.33      \\
    -Quant &1.61 & 9.67& 13.51\\
    \bottomrule
  \end{tabular}
\end{table}

\subsection{Resources Requirement}

Due to quantization preprocessing, our \textbf{PTQ\textit{1.61}} has higher resource cost compared to other PTQ methods. For example, compared to OmniQuant, as shown in Table \ref{cost}, our method has slightly higher memory cost.
Although this is a limitation of our method, it remains within an acceptable range (compared with OneBit which needs 5 A800 GPUs and over 24days to train a low-bit LLaMA-7B) considering we target the most challenging extremely low-bit scenario (it is really hard to ensure performance especially for PTQ), and the performance is better than OmniQuant which is current SOTA in extremely low-bit PTQ even without quantization preprocessing as indicated in Table \ref{appendix-ppl}.

\begin{figure*}[t]
  \centering
  \includegraphics[width=5.4in]{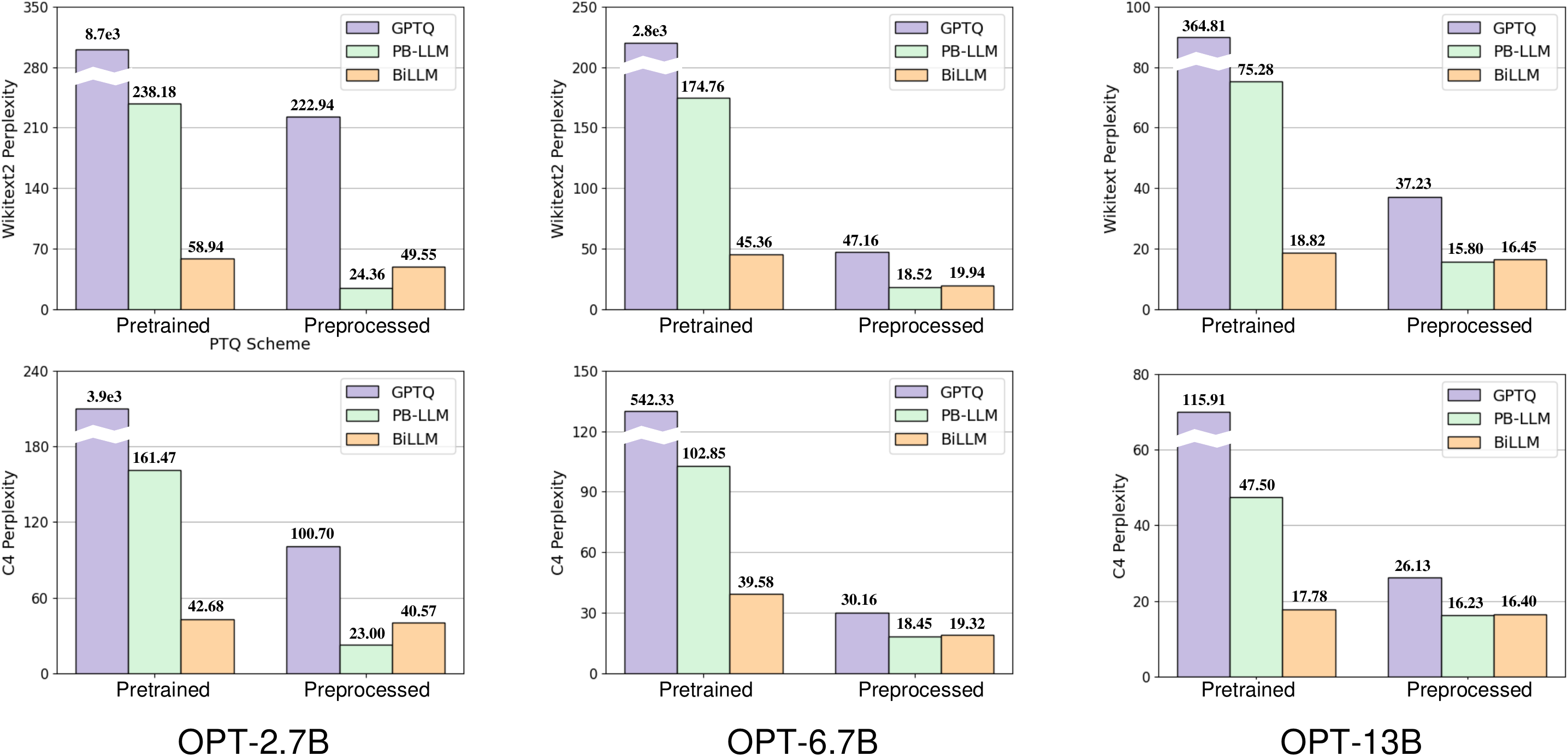}%
  \caption{Our novel quantization preprocessing scheme on OPT families quantized other existing PTQ methods.}
  \label{bar-chart-opt}
\end{figure*}

\subsection{OPT Results on Other Approaches}
\label{opt-other}
The effectiveness of proposed quantization preprocessing scheme on LLaMA families quantized by other existing low-bit PTQ methods has been illustrated in Figure \ref{bar-chart}. Besides, results on OPT families is available in Figure \ref{bar-chart-opt} and the similar phenomenon can be observed.

\begin{figure*}[t]
  \centering
  \includegraphics[width=4.5in]{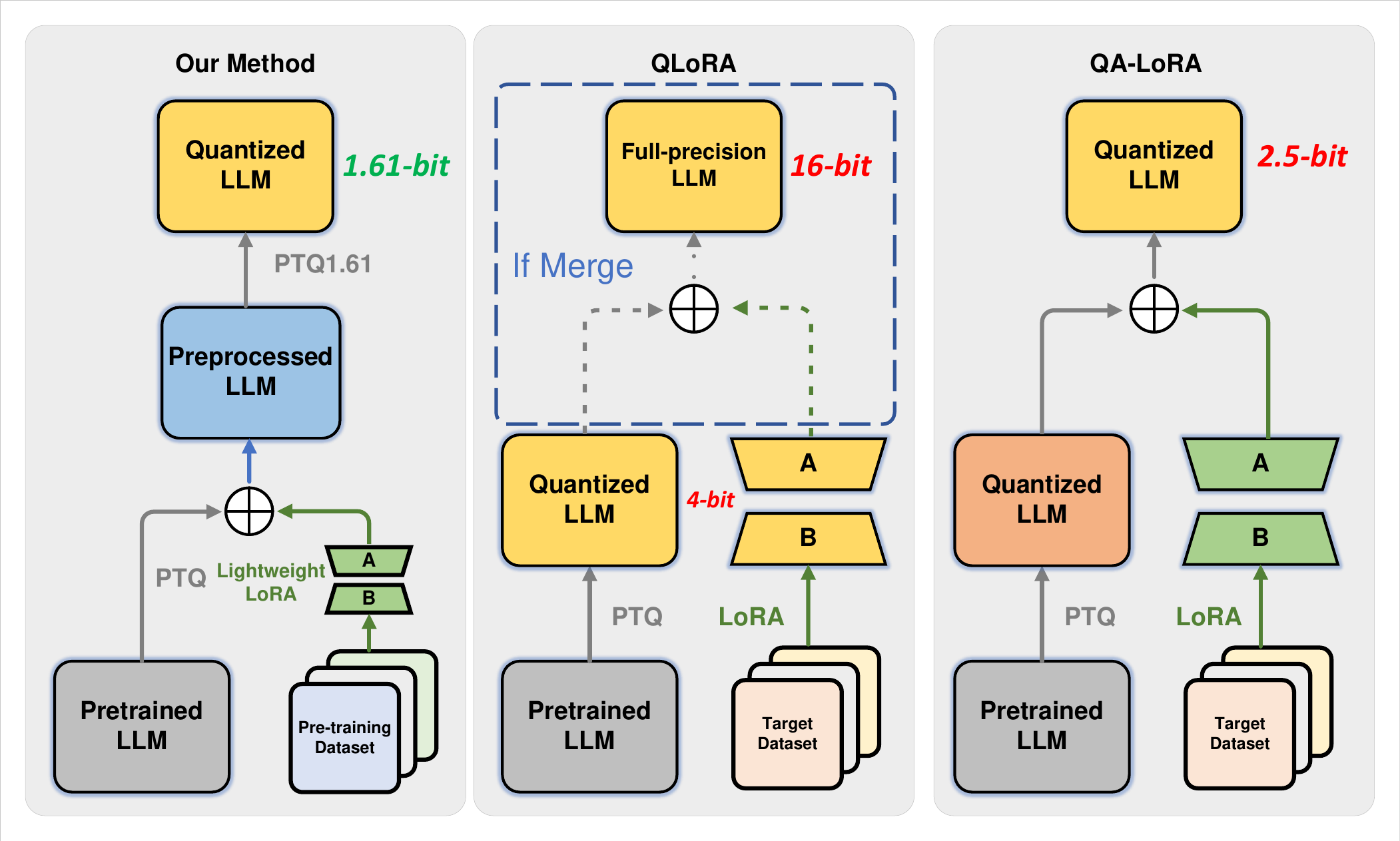}%
  \caption{Comparison between our method and post-quantization PEFT methods. The yellow part of each method will be actually deployed and loaded during inference.}
  \label{lora}
\end{figure*}

\begin{table}
\small
\centering
    \begin{tabular}{ccc}
    \toprule
    \multirow{2}{*}{Model} & \multicolumn{2}{c}{Dataset}       \\
    \cmidrule(l){2-3}
       & WikiText2     & C4 \\
    \midrule
    LLaMA-7B & 287.38  & 829.91     \\
    LLaMA-2-7B & NAN & NAN      \\
    \bottomrule
  \end{tabular}
  \caption{Perplexities on LLMs when setting learnable row-wise mean which is similar to zero-point in QA-LoRA (group-size=1) for binarization.}
\label{qalora-ppl}
\end{table}

\subsection{Comparison with Post-Quantization PEFT Methods}
\label{peft}

It is worth mentioning that the restorative LoRA in our quantization preprocessing scheme stands out significant advantages and differences from existing post-quantization PEFT methods \citep{dettmers2023qlora,xu2023qa}, as shown in Figure \ref{lora}. For advantages: \textbf{(a)} Compared with LoRA and QLoRA \citep{dettmers2023qlora}: Both methods requires to store additional float-point low-rank matrices, augmenting inference costs. Merging these matrices with the low-bit quantized model will reinstate it to FP16, imposing considerable storage demands. Conversely, our preprocessing approach concurrently optimizes storage and inference. \textbf{(b)} Compared with QA-LoRA \citep{xu2023qa}: QA-LoRA addresses the aforementioned issues by adjusting the group-wise zero-point in Equation \eqref{higher-bit}, which is similar to the mean used in binarization. Specifically, QA-LoRA partitions each row of the weight matrix into numerous groups (termed group-size) and determines a distinct zero-point for each group, thereby circumventing the need for additional memory during fine-tuning. However, their fine-grained group-size of zero-point is set to 32, which introduces an additional 0.5-bit per weight, leading to extra storage overheads. Our experiments reveal that setting the group-size to 1 in QA-LoRA as us to learn a row-wise mean significantly compromises performance, as demonstrated in Table \ref{qalora-ppl}. 

The remaining differences can be summarized as: \textbf{(a)} Objective: PEFT aims to enhance model's performance on downstream tasks, whereas restorative LoRA targets to transform salient weights to a concentrated row-wise pattern for better quantization. \textbf{(b)} Datasets: PEFT utilizes datasets from the target domain, such as Alpaca \citep{alpaca} and FLAN v2 \citep{longpre2023flan}, while restorative LoRA tends to leverage the pre-training datasets, \textit{i.e.}, RedPajama \citep{together2023redpajama}. \textbf{(c)} Time cost: For LLaMA-7B, PEFT usually takes over 10+ hours (even the less QA-LoRA requires over 6 hours), whereas our lightweight restorative LoRA is cost-friendly, requiring less than 1.2 hours.

\subsection{Comparison with QAT}
\label{qat}

\begin{figure*}[t]
  \centering
  \includegraphics[width=6.2in]{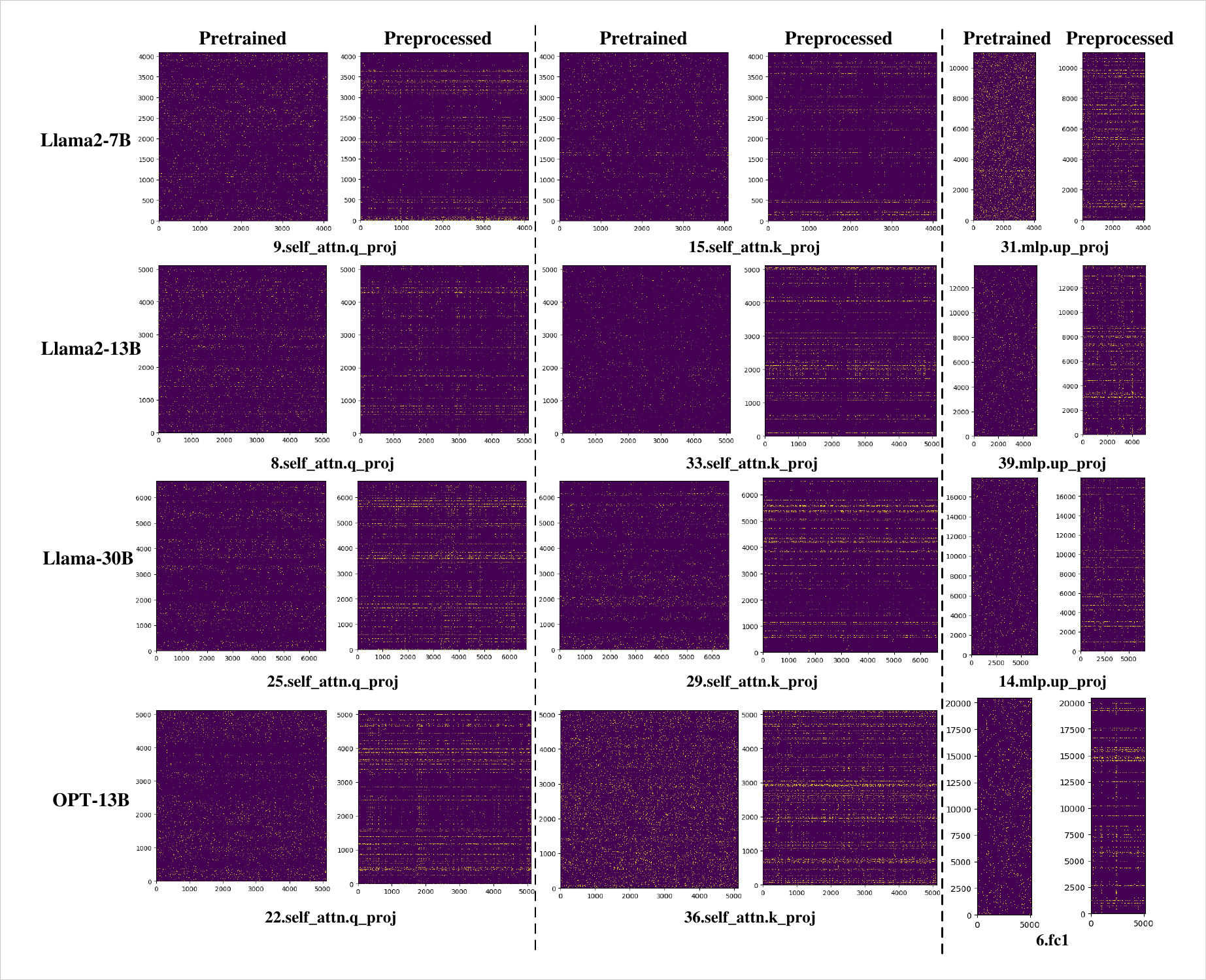}%
  \caption{More visualizations which show the impact of our quantization preprocessing method.}
  \label{more visual}
\end{figure*}

As is known to all that QAT frameworks improve quantization performance by directly training a quantized LLM to get optimal weights as well as quantization parameters to accommodate quantization errors at target bit-width, where its strategy for adjusting weights may share some similarities with us. To eliminate this confusion, several important distinctions and limitations need to be highlighted: \textbf{(a)} QAT trains all the weights in the LLM, which incurs extremely high training costs, \textit{i.e.}, LLM-QAT \citep{liu2023llm} requires 3 days to retrain a quantized OPT-1.3B on 8 Nvdia-A100 GPUs and OneBit \citep{xu2024onebit} spends over 24 days to retrain a LLaMA-7B on 5 Nvdia-A800 GPUs. In contrast, our quantization preprocessing only requires to transform a few channels into a PTQ-friendly row-wise format utilizing the low-rank nature of weight compensation, with only a single Nvdia-A100 GPUs for less than 1.2 hours. \textbf{(b)} QAT needs to know the target bit-width at the beginning and necessitates retraining from scratch if the bit-width changes, making it inflexible. While our method only requires saving the preprocessed model, which can then be quickly adapted to any target bit-width because the row-wise pattern is universally applicable to per-channel PTQ across various target bit-widths. \textbf{(c)} It is well-known that training a binary quantized LLM with gradient descent is highly challenging \citep{qin2020binary}. Specifically, QAT introduces the straight-through estimator to approximate quantization gradients. However, under extremely low-bit conditions, the gradient approximation error becomes significantly large, making the training process more unstable and harder to converge. By comparison, our preprocessing strategy naturally addresses this by placing the restorative LoRA process before quantization, thereby avoiding the associated optimization difficulties.

\subsection{More Visualizations}
For a deeper understanding of the quantization preprocessing impacts on salient weights distribution, we provide more comprehensive visual data, including a variety of layers and models as Figure \ref{more visual}. The similar channel-wise nature occurs in various preprocessed models and layers which demonstrates that our quantization preprocessing process is essential.


\section{More Evaluations}
\label{more-eval}
\subsection{MMLU and GSM8K}
In addition to the benchmarks in the content, we also valid the quantization performance on GSM8K and MMLU on several LLMs. However, as the near-random levels illustrated in Table \ref{mmlu/gsm8k}, we observe that under extremely low-bit quantization, all existing PTQ methods nearly make LLMs loss the ability, which is consistent with previous research \citep{liu2023emergent}. Considering the disappointing outcomes, we choose not to list the comparison into the content.

\subsection{Long Context Understanding}
LongBench is a novel benchmark for evaluating long context understanding capability which is a critical measurement for LLMs application. Due to LongBench only supports Chat-LLMs, we select LLaMA2-7b-Chat for evaluation in Table \ref{longbench}. A consistent superior performance proves the effectiveness of our method.

\begin{table}[t]
  \small
  \centering
  \begin{tabular}{cccc}
    \toprule
    \textbf{LLaMA}&\textbf{Method} &\textbf{MMLU} & \textbf{GSM8K}\\
    \midrule
    \multirow{3}{*}{1-7B} &PB-LLM &23.0&0.23\\
     &BiLLM &22.9& 0\\
     &\textbf{PTQ\textit{1.61}} &23.0&0.15\\
    \midrule
    \multirow{3}{*}{2-7B}&PB-LLM &23.0&0\\
     &BiLLM &22.9& 0\\
     &\textbf{PTQ\textit{1.61}} &22.9&0.61\\
     \midrule
    \multirow{3}{*}{3-8B}&PB-LLM &22.9&0.83\\
     &BiLLM &22.8& 0.30\\
     &\textbf{PTQ\textit{1.61}} &22.9&0.83\\
    \bottomrule
  \end{tabular}
    \caption{Evaluation accuracies on MMLU and GSM8K.}
  \label{mmlu/gsm8k}
\end{table}

\begin{table}[t]
  \small
  \centering
  \begin{tabular}{cccc}
    \toprule
    \textbf{LongBench} &PBLLM&BiLLM &\textbf{PTQ\textit{1.61}}\\
    \midrule
    2WikiMQA & 5.0 &4.1 &\textbf{12.7}  \\
    TriviaQA &  9.9 &13.2 &\textbf{34.8}  \\
    Multi-News & 13.6 &11.7 &\textbf{14.6}  \\
    SAMSum & 3.3 &5.4 &\textbf{19.5}   \\
    QMSum & 5.4 &8.3 &\textbf{15.6}   \\
    MultiFieldQA-EN & 8.4 &8.1 &\textbf{13.2}   \\
    \bottomrule
  \end{tabular}
    \caption{Evaluation on LongBench.}
  \label{longbench}
\end{table}

\begin{table}[t]
\small
  \centering
  \begin{tabular}{ccc}
    \toprule
    \textbf{Method} &LLaMA-1/2-7B & LLaMA-1/2-13B \\
    \midrule
    PB-LLM &2.36GB&4.49GB\\
    BiLLM &1.83GB& 3.50GB\\
    \textbf{PTQ\textit{1.61}} &\textbf{1.41GB}& \textbf{2.68GB}\\
    \bottomrule
  \end{tabular}
    \caption{Inference memory comparison.}
  \label{inference}
\end{table}

\subsection{Throughput and Inference Memory}
For real-world system evaluation, current NVIDIA GPUs do not yet support such low-bit inference. Designing specific hardware and operation kernals that meets the inference conditions requires larger research teams and financial support, so our goal is to explore the performance limits of PTQ by fake-quantization before commercial hardware support is available. We believe this will eventually be realized, as evidenced by the latest NVIDIA GPUs now supporting 4-bit inference, whereas only a year ago they were limited to 8-bit. 

Compared with PB-LLM and BiLLM which requires to load extra unstructured mask during inference, our method is much more efficient. Followed by previous research \citep{ma2024era}, we can obtain information about a 1.58-bit (ours is 1.61-bit) LLaMA-7B achieves a 2.9X speedup in latency and LLaMA2-70B gains an 8.9X increase in throughput (2977 tokens/s). 

In addition, we provide the memory usage of LLMs quantized by PB-LLM, BiLLM and \textbf{PTQ\textit{1.61}} via calculation considering weight bits, scaling factors and masks. As indicated by Table \ref{inference}, our \textbf{PTQ\textit{1.61}} has an advantage in memory efficiency, which is of practical benefits.

\section{Discussion on Practically Applicability of Extremely Low-bit Weight Quantization}

\subsection{Accuracy-Latency Tradeoff Analysis}
One important aspect of LLM quantization is the accuracy-latency tradeoff. If speed is the sole priority, an aggressive compression ratio can significantly improve latency, but this often results in unacceptable accuracy degradation. Conversely, adding additional components can help recover accuracy but may hinder system acceleration, leading to slower performance. Therefore, it is crucial to analyze the accuracy-latency tradeoff.

Delve deeper into the calculation process of a quantized model on a GPU. For weight-only quantization, the primary acceleration comes from the transfer of low-bit integer weights from memory to the MAC (Multiply Accumulate) processing unit, which reduces the amount of data transfer compared with FP model. To restore performance, existing quantization methods introduce additional components, such as FP16 channel-wise scaling factors, to reduce quantization errors. For the attention layer of an LLM, the size of the weight matrix is usually $4096\times4096$, and the scaling factors are a $1\times4096$ vector. Therefore, in the transfer process mentioned in the previous paragraph, the bad impact on latency from transferring such a small amount of FP16 scaling factors is almost negligible compared to the significant inference acceleration benefits brought by transferring the low-bit weight matrices. Considering all above, a small amount of additional components such as scaling factors will not have a significant bad impact on the inference latency in model quantization.

\subsection{Compared with Weight-activation Quantization}
In order to prove the necessity of extremely low-bit weight PTQ research, we provide the performance gap among FP16 results, W4A4 SmoothQuant and extreme low-bit PTQ methods. The results are shown as Table \ref{discuss}. The results demonstrate the necessity of research into extremely low-bit PTQ from two aspects. Firstly, compared to the results of FP16, previous method (PBLLM) indeed showed a significant gap, but our PTQ1.61 has narrowed the performance gap to an acceptable level. Secondly, compared to SmoothQuant, the most popular and wide applied weight-activation PTQ method, its performance of W4A4, which is currently supported by the latest commercial GPUs, is still inferior to our PTQ1.61, proving the research prospect of extremely low-bit weight quantization is as bright as weight-activation quantization. We are confident that with further advancements, the disparity between extremely low-bit weight quantization and full precision will progressively diminish.

\begin{table}[t]
\tiny
  \centering
  \setlength{\tabcolsep}{3pt}
  \begin{tabular}{ccccccccc}
    \toprule
    \textbf{Model} &\textbf{Method} &PIQA & ARC-e &ARC-c &HellaS &WinoG &Race & \textbf{Avg.} \\
    \midrule
    \multirow{4}{*}{LLaMA-13B}&FP &79.16 &77.31 &46.42 &59.90 &72.93 &39.71 &62.57\\
    &PB-LLM &60.55 &37.46 &18.69 &30.79 &51.07 &30.24 &42.75\\
    &SQ(W4A4)&62.45 &44.31 &24.48 &35.63 &50.11 &31.74 &41.45\\
    &\textbf{PTQ\textit{1.61}} &\textbf{68.17} &\textbf{58.59} &\textbf{27.22} &\textbf{40.02} &\textbf{58.33} &\textbf{34.26} &\textbf{47.77}\\
    \midrule
    \multirow{4}{*}{LLaMA-30B}&FP &80.96 &80.39 &52.73 &63.34 &75.69 &40.57 &65.61\\
    &PB-LLM &64.91 &46.38 &21.33 &35.81 &61.17 &30.91 &43.42\\
    &SQ(W4A4)&54.57 &28.82 &19.45 &26.93 &49.88 &22.52 &33.70\\
    &\textbf{PTQ\textit{1.61}} &\textbf{70.24} &\textbf{63.64} &\textbf{32.17} &\textbf{46.82} &\textbf{63.61} &\textbf{37.13} &\textbf{52.27}\\
    \midrule
    \multirow{4}{*}{LLaMA2-13B}&FP &79.05 &79.38 &48.38 &60.04 &72.14 &40.48 &63.25\\
    &PB-LLM &54.46 &27.95 &19.54 &26.74 &49.96 &26.03 &34.11\\
    &SQ(W4A4)&54.18 &28.78 &19.28 &27.82 &50.19 &27.26 &34.59\\
    &\textbf{PTQ\textit{1.61}} &\textbf{66.54} &\textbf{56.86} &\textbf{26.45} &\textbf{40.32} &\textbf{55.88} &\textbf{33.30} &\textbf{46.59}\\
    \bottomrule
  \end{tabular}
    \caption{Performance comparison among FP16, SmoothQuant(W4A4), PB-LLM and our \textbf{PTQ\textit{1.61}}.}
  \label{discuss}
\end{table}

\end{document}